\documentclass{article}

\usepackage[preprint]{neurips_2026}

\usepackage[utf8]{inputenc} 
\usepackage[T1]{fontenc}    
\usepackage{hyperref}       
\usepackage{url}            
\usepackage{booktabs}       
\usepackage{amsfonts}       
\usepackage{nicefrac}       
\usepackage{microtype}      
\usepackage{xcolor}         
\usepackage{booktabs}
\usepackage{multirow}
\usepackage{xspace}
\usepackage{graphicx}
\usepackage{cleveref}
\usepackage{arydshln}
\usepackage{enumitem}

\title{Implicit Rule Induction with Test-Time Task Embeddings in \ARClike Tasks}

\newcommand{\ARClike}{ARC-like\xspace}
\newcommand{\ourTTT}{Embed-TTT\xspace}
\newcommand{\embedfullTTT}{Embed-Full-TTT\xspace}
\newcommand{\defaultTTT}{Full-TTT\xspace}

\author{%
  Adrien~Deliège\thanks{Equal contribution. Correspondence to \texttt{adrien.deliege@uliege.be} and \texttt{claasbeger@santafe.edu}} \\
  University of Liège\\
   \And
   Claas Beger\footnotemark[1] \\
   Santa Fe Institute \\
   \AND
   Marc Van Droogenbroeck \\
   University of Liège \\
   \And
   Melanie Mitchell \\
   Santa Fe Institute \\
}

\begin{document}

\maketitle

\begin{abstract}
The Abstraction and Reasoning Corpus and related benchmarks evaluate whether AI models can solve novel reasoning tasks, but often leave unclear whether success reflects inference of the intended underlying rule or reliance on shortcuts. We address this gap by studying test-time task embeddings in Vision ARC (VARC), a model in which a pre-trained backbone is complemented by a trainable embedding representing the transformation rule. In the original VARC, test-time training (TTT) is jointly applied to  the backbone and task embedding.  Here we introduce a novel two-step TTT protocol: first finetune only the task embedding (\ourTTT), then freeze it and finetune the backbone. Across ARC-AGI-1, ConceptARC, and two controlled datasets with known rules, \ourTTT consistently yields improved task embeddings, ones that align better with underlying task rules, improve embedding-based retrieval, and enable accurate linear probing of known rules. Qualitatively, \ourTTT identifies more semantically meaningful relations between test and train tasks on ARC-AGI-1. We also show that optimizing only task embeddings (less than 0.01\% of model parameters) already solves a non-trivial fraction of ARC-AGI-1, ConceptARC and Mini-ARC tasks, while the full two-step pipeline improves final performance. Finally, we show that \ourTTT recovers the underlying geometric structure of parametric rules and learns compositional capabilities that enable rule-wise interpolation, but not extrapolation. These findings support a clearer separation between rule induction and rule execution in ARC-like evaluations, motivating benchmarks that better distinguish in-distribution from out-of-distribution rules.
\end{abstract}

\section{Introduction}

The Abstraction and Reasoning Corpus (ARC) was introduced to evaluate a form of humanlike fluid intelligence that is difficult to capture with standard machine learning benchmarks~\citep{Chollet2019OnThe-arxiv}. In each task, a model is given only a few input-output grid demonstrations and must infer the transformation rule that maps inputs to outputs. This setup has inspired a growing family of \ARClike benchmarks for studying humanlike abstraction, few-shot adaptation, and systematic generalization. However, standard \ARClike evaluations typically measure whether a model produces the correct output, not whether it has inferred the intended rule. This distinction is important: a model may solve a task by inducing the underlying rule, but it may also exploit dataset-specific shortcuts, memorize related patterns, or rely on test-time optimization without forming a reusable representation of the task~\citep{Beger2025DoAI-arxiv}. Conversely, a model may identify a useful rule but fail to execute it perfectly. As a result, task accuracy alone conflates rule induction, execution and memorization.
This issue is amplified by the fact that most \ARClike datasets do not provide the rules or generators underlying their tasks. Consequently, it is usually unclear whether rules underlying test tasks are in or out-of-distribution with respect to the rules known from training tasks, where training tasks would be drawn from the ARC-AGI-1 training set and test tasks from ARC-AGI-1 evaluation set, for example. From a classical machine learning perspective, this makes benchmark results of models trained on the training tasks difficult to interpret: strong performance may reflect genuine abstraction, but it may also reflect favorable overlap between train and test rules. We therefore argue that \ARClike datasets and evaluation should more explicitly study whether models induce meaningful task representations, and should distinguish rule induction from rule execution whenever possible. 

In this paper, we study implicit rule induction through the task embeddings of Vision ARC (VARC)~\citep{Hu2026ARCIs}, a ViT-based model that represents each task with a trainable embedding, which steers the output grid generation in a separate model backbone via a latent representation of the underlying rule. We show that the default VARC test-time training procedure, which jointly finetunes the task embedding and the model backbone, produces poorly aligned test-train task embeddings that do not retain embedding space information from the training stage.
We propose a simple alternative, \ourTTT: at test time, we freeze the backbone and optimize only the task embedding. This forces test tasks to be represented in the same embedding space as training tasks, as both pass through the same frozen backbone, making the resulting embeddings more suitable for analyzing whether the model has identified the underlying rule. We evaluate these embeddings on ARC-AGI-1 (written ARC-1 for brevity)~\citep{Chollet2019OnThe-arxiv}, ConceptARC~\citep{Moskvichev2023TheConceptARC}, and two controlled \ARClike datasets with known rules. When ground-truth rules are unavailable, we test whether duplicated training tasks recover their original embeddings. When rules are available, we evaluate whether ground-truth rule labels or parameters can be decoded from embeddings using linear probes. Across these settings, \ourTTT yields substantially more rule-aligned embeddings than default full-model test-time training, while also supporting increased task-solving performance when followed by optional backbone finetuning. Overall, we make the following \textbf{contributions}:

\begin{itemize}[leftmargin=*, itemsep=2pt, topsep=2pt]
    \item We propose a novel test-time training protocol, \ourTTT, for VARC, which produces semantically meaningful task embeddings for \ARClike tasks. To evaluate \ourTTT against the default full model test-time training, we further introduce an evaluation methodology for implicit rule induction.
    \item We show quantitatively that \ourTTT embeddings align better with task rules across four \ARClike datasets, and we provide qualitative evidence of meaningful test--train semantic similarity in ARC-1. In addition, optimizing only task embeddings solves 17\% of ARC-1 test tasks, and nearest-neighbor concept transfer from ConceptARC to ARC-1 yields 36\% correct concept associations, without training for that transfer objective nor for concept classification.
    \item We analyze the task embedding space on a specific case and characterize which unseen tasks remain solvable. We show that task embeddings recover the dataset geometry and support compositional rule interpolation, reframing \ARClike performance through a standard in-/out-of-distribution lens.

\end{itemize}

\section{Related work}

\subsection{Problem statement} 

\textbf{Definitions.} An \emph{\ARClike dataset} is a set of \emph{tasks}, and a task is a set of $(x,y)$ pairs governed by a deterministic \emph{rule} that maps each \emph{input} $x$ to its \emph{output} $y$. In theory, the rule can range from a simple function to a complex algorithm, but implicitly it is commonly accepted that the rule should be expressible in natural language and understandable by human readers. Following the original ARC-1 format, each $x$ and each $y$ is a two-dimensional grid of integers, such that a task can be viewed as a visual puzzle, where each cell of the grid is colored according to its integer value. Typically, each task is split into demonstration pairs $(x_i,y_i)$ and a test input $x'$, such that a solver should infer a common rule $R$ that maps each $x_i$ to $y_i$ and then should apply $R$ to map the test input $x'$ to $y_i$. To elicit in-context test-time learning, an \ARClike dataset is usually split into training tasks and test tasks, such that one should learn generic task-solving strategies on the training tasks, and successfully apply them on the test tasks without any human intervention. \newline
\textbf{Performance.} The performance of a model on an \ARClike dataset is the proportion of test tasks solved. A test task is solved when the test input $x'$ is correctly mapped to its ground-truth output $y'$, which means reaching 100\% cell-wise accuracy on the $y'$. In practice, several predictions can be made for a given test input $x'$, and the task is considered solved if one of two predictions matches the ground truth (``pass @ 2''). In general for \ARClike benchmarks, while ground-truth outputs $y'$ are given for each task, the explicit rules underlying the tasks are not provided, which makes it difficult to evaluate whether the model actually learned the intended logic behind the tasks. 

\subsection{Approaches}

\textbf{Implicit and explicit reasoning.} Several works have pointed out the tendency of large language models (LLMs) to rely on shortcuts or shallow reasoning on tasks requiring analogy or abstraction \citep{Beger2025DoAI-arxiv, Mineault2026Cognitive-arxiv, Lewis2025Evaluating}. Many benchmarks targeting these abilities are susceptible to such behavior, as they primarily evaluate the correctness of the final output rather than the actual reasoning process. Recent work has therefore shifted toward learning more robust representations, either internally through structured latent models such as World Models \citep{Piriyakulkij2025PoEWorld, Maasch2025CausalARC}, or externally via explicit reasoning artifacts, for instance in program synthesis approaches \citep{Langenfeld2026Bongards-arxiv, Li2025Combining}. In this work, we aim to bridge these perspectives by enforcing the model to store its internal representation in an explicit vector that can be directly extracted and ``executed,'' for instance through linear projections.\newline
\textbf{Embedding vectors.} The extraction of embedding or activation vectors is a common technique in interpretability research. \citet{Todd2024Function} identify sets of attention heads that implement simple functions, such as word translation or antonym mapping, and show that combining such function vectors yields non-trivial compositions of functionality. We extend this line of work to models applied to abstract reasoning tasks by introducing a test-time tuning regime that encourages the formation of explicit task embeddings. We analyze the extent to which these embeddings capture underlying rules, including their compositional properties and limitations. We also study similarities between embeddings of tasks that share related rules. Related observations, albeit in a more correlational setting, have been reported by \citet{Lake2018Generalization, Hill2019Learning, Veldkamp2023Solving}.\newline
\textbf{Test-time adaptation in abstract reasoning tasks.} Test-time training (TTT) has been used in several high-performing systems for solving ARC-1 tasks \citep{Li2025Combining, Chollet2024ARCPrize-arxiv, Hu2026ARCIs,Akyurek2025TheSurprising}. TTT typically works by generating additional training examples through simple augmentations at test time. However, while TTT improves model performances on ARC-1, it remains unclear whether TTT leads to the induction of meaningful task representations or instead relies on unconstrained parameter updates. One of the motivations of our work is to study this distinction, by constraining TTT to operate primarily through explicit task embeddings rather than full model finetuning.\newline 
\textbf{Vision ARC (VARC).} The current best-performing models on \ARClike datasets are closed-source LLMs, from which internal representations cannot be extracted. A rare exception is the Vision ARC (VARC) model~\citep{Hu2026ARCIs}, which is a ViT-based model trained from scratch that explicitly models each task through a specific task token, numerically encoded as a $512$-dimensional array called a \emph{task embedding}. During training, the model learns to solve training tasks as image-to-image translation problems, with each problem conditioned by its corresponding trainable task embedding. Then, for a given test task, the model randomly initializes the task embedding and follows a TTT pipeline, which learns a new task embedding jointly with finetuning the trained backbone (without the task embedding part) on the test task at hand. We call this process the baseline \defaultTTT approach.

\section{Method}

\subsection{Obtaining rule-inductive test-time task embeddings}


\textbf{Issues with test embeddings.} The original VARC paper ~\citep{Hu2026ARCIs} shows that \emph{train} task embeddings seemingly cluster semantically similar tasks and shared (sub-)rules, suggesting that embeddings can support implicit rule induction. However, \emph{test} embeddings produced by \defaultTTT were not analyzed. In our preliminary experiments, test embeddings exhibit a strong distribution shift from train embeddings (in both direction and magnitude) and often stay close to their random initialization. As a result, semantic neighborhood structure is largely lost, weakening rule-induction usefulness at test time. Although this mismatch does not drastically hurt final task-solving performance, it raises efficiency and interpretability concerns: even for duplicates of training tasks presented at test time, the pipeline finetunes 18M parameters per task yet fails to recover the corresponding train embeddings. Moreover, we observed that freezing the task embedding (random or zero initialization) causes little performance drop, indicating that the backbone can ignore it. We argue that a desirable model should produce consistent, rule-inductive task embeddings in both training and test phases.

\begin{figure}[t]
  \centering
  \includegraphics[width=\linewidth]{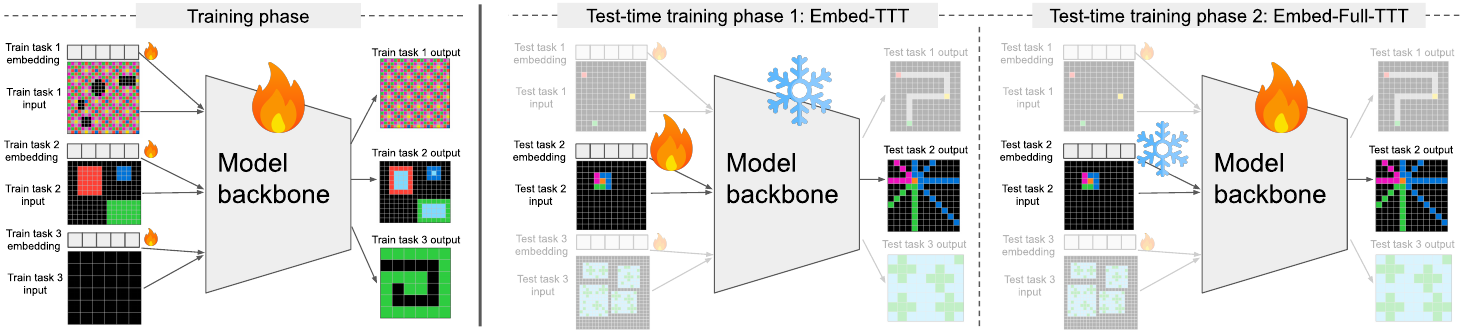}
  \caption{\textbf{Method.} After training a model that learns task embeddings that condition the backbone (left), we elicit implicit rule induction at test time: we first freeze the backbone and learn a task embedding (\ourTTT, center), then freeze the embedding and finetune the backbone (right).}
  \label{fig:method}
\end{figure}

\textbf{A two-step finetuning approach.} Drawing on the previous observations, we derive a TTT protocol that naturally allows learning of meaningful test task embeddings while also improving performance capabilities. It consists of a two-step finetuning approach, shown in~\Cref{fig:method}. When faced with a test task, we first freeze the model backbone and learn only the task embedding. We denote this step as \ourTTT. We then freeze the learned task embedding and finetune the model backbone (\embedfullTTT). \ourTTT enables positioning the test task embedding in the same embedding space as the training task embeddings because they share the same model backbone, while \embedfullTTT allows the model to adjust to the learned embedding to solve the task efficiently.


\subsection{Evaluating implicit rule induction through task embeddings}
\label{sec:eval_rule_induc}

\textbf{Rules unavailable: training embeddings retrieval.} In datasets like ARC, where each task is unique, automated evaluation of rule induction largely reduces to checking whether duplicate training tasks retrieve their originally learned embeddings at test time. 
More precisely, we compute the proportion of tasks for which the original embedding is within the top-k closest embeddings of the newly TTT-produced embedding. Limitations of this approach are the fact that memorization alone might suffice to retrieve the original task embedding, the fact that multiple embeddings might solve the tasks, and the fact that, for training tasks that the model was not able to solve, different embeddings might be better suited than the original one. 
Consequently, this default protocol should be interpreted with caution: it provides only indirect evidence of meaningful implicit rule induction at test time, mostly reliable when test rules are in-distribution with respect to the training rules distribution.

\textbf{Rules available: linear probing to rules.} When rules (or concept groups) are defined and task generators can produce multiple task instances per rule, in-distribution rule induction can be evaluated more cleanly. The setup mirrors self-supervised learning: learn training embeddings through an auxiliary objective, then assess test embeddings with a linear probe on a downstream target. In our \ARClike setting, each data ``unit'' is a task, and the auxiliary objective is solving the task by finetuning its task embedding. Given rule labels for training tasks, we collect train embeddings, fit a linear probe from embeddings to rules, and apply it to test embeddings. We probe performance then measure how much rule information is encoded in the embeddings. 
We distinguish two cases:
\begin{itemize}[leftmargin=*, nosep]
    \item \textbf{Non-parametric rules.} Linear probing is a classification problem, and we measure implicit rule induction by the accuracy on the test tasks of the linear classifier learned from the train tasks. 
    \item \textbf{Parametric rules.} Linear probing is a linear regression problem, and we measure the root mean squared error (RMSE) between linearly regressed and ground-truth parameters on the test tasks from the linear regressor learned from the train tasks.
\end{itemize}

\clearpage

\section{Results}
\subsection{Quantitative in-distribution implicit rule induction evaluation}

\textbf{Datasets and experimental details.} We conduct the following experiments.

\begin{itemize}[leftmargin=*, nosep]
    \item ARC-1. We start from the trained VARC of \citet{Hu2026ARCIs} and perform \defaultTTT and \ourTTT on duplicates of the training tasks to measure train embedding retrieval accuracy.
    \item ConceptARC. We train a VARC model from scratch and perform the same experiment as for ARC-1. 
    We also evaluate the linear probing classification accuracy by considering the 16 concepts underlying the tasks as proxies for the rules that generated them. Since no train-test split exists, we perform a 5-fold cross-validation, each time learning the linear mapping from 8 tasks per concept and evaluating it on the remaining 2 tasks. We report the average accuracy across the runs. 
    \item Custom ``LUCD''. We design a simple dataset with known rules in the \emph{non-parametric} setting. Each task consists in applying zero, one, or two transforms to a single-cell object positioned randomly on a $10 \times 10$ grid, sampled (with repetition enabled) from 4 ``atomic'' transforms: move one cell to the left (L), up (U), dilate in the cardinal directions (C), dilate in all directions (D). Since all atomic rules commute with each other, this yields 15 different possible rules. 
    \item Custom ``Moves''. We design a simple dataset with known rules in the \emph{parametric} setting. Each task consists in moving a single-cell object on a $10 \times 10$ grid by at most five cells both horizontally (left (L), right (R)) and vertically (up (U), down (D)), yielding a total of 121 different rules (for instance, ``L4-D3'' is the rule moving 4 cells to the left and 3 down, associated to coordinates (-4,-3)). 
\end{itemize}
For ``LUCD'' (resp. ``Moves''), we generate 1 training task and 10 test tasks per rule, with 10 training and 10 test input-output pairs per task(we provide examples in~\autoref{sec:LUCD} (resp.~\autoref{sec:moves})). We evaluate rule induction capabilities on these datasets as described in~\Cref{sec:eval_rule_induc}.


\textbf{Results.} \Cref{tab:perf-results} reports the results of the experiments. Across all evaluations, \ourTTT largely outperforms the baseline \defaultTTT, which barely outperforms pure chance of retrieving or classifying the embeddings correctly. Distances defining neighbor ordering of embeddings are computed as cosine similarity distances. Also, we observed that tasks successfully solved are generally better retrieved (91.2\% with \ourTTT on ARC) than failed tasks (77.1\%). In the case of ConceptARC, it should be noted that the concepts are only a meta-category associated with the tasks and do not fully characterize the task-generating process. Therefore, observing a perfect mapped accuracy is likely out of reach, but \ourTTT performs reasonably well, with 60.6\% accuracy (chance is 6.2\%). In our custom LUCD dataset, one-shot learning a linear classifier from a single task instance per rule yields a perfect test accuracy, whereas \defaultTTT performs at chance. This is a strong sign that \ourTTT is capable of implicit rule induction, since test tasks were never seen during training (in contrast with the previous evaluations). We observe the same trend for our regression evaluation on the Moves dataset. As an additional variant, we can align the probing setup with each method's own test-time dynamics: we first run each TTT protocol on duplicates of the training tasks, then learn the linear mapping from these TTT-produced train embeddings (instead of from frozen post-training embeddings), and finally apply it to test embeddings. Under this setting, on LUCD, \defaultTTT rises to 39.2\% while \ourTTT remains at 100\%; on Moves, \defaultTTT improves to 1.27 of RMSE and \ourTTT reaches 0.44 of RMSE.

\begin{figure*}[t]
  \centering
  \includegraphics[width=0.495\textwidth]{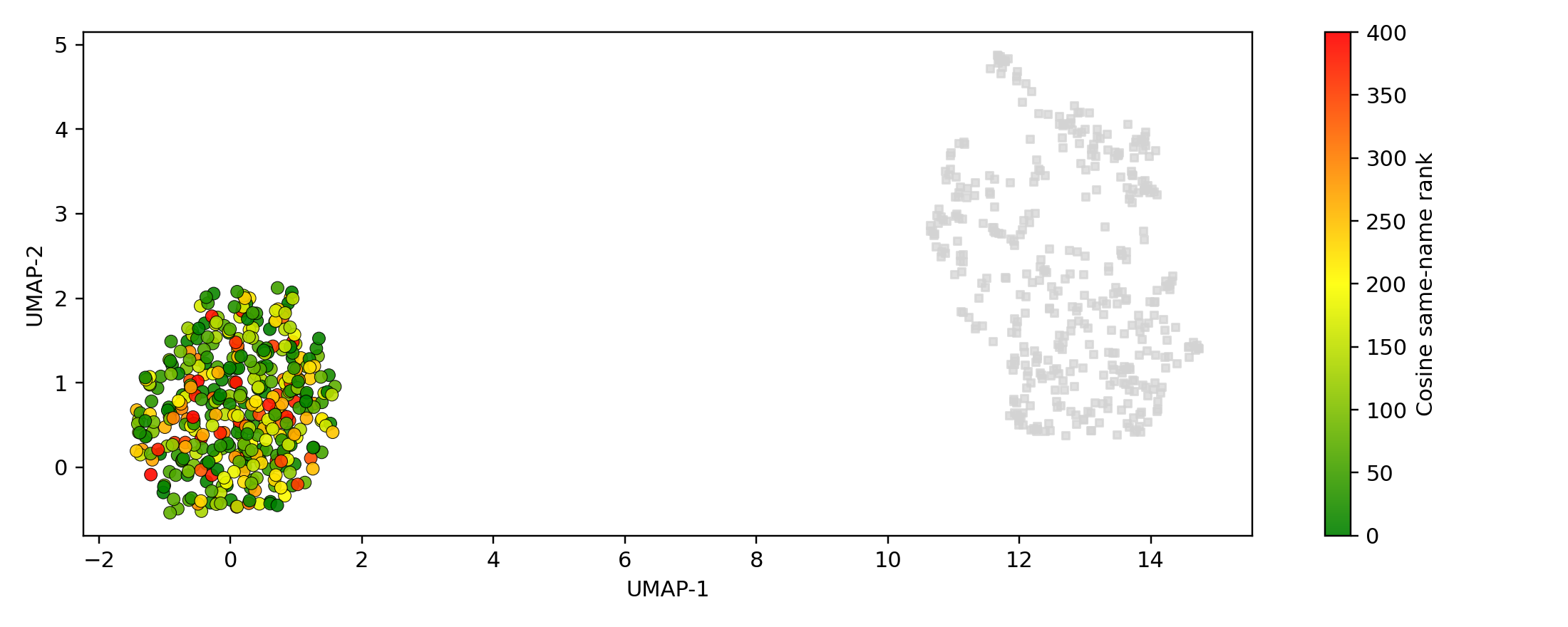}\hfill
  \includegraphics[width=0.495\textwidth]{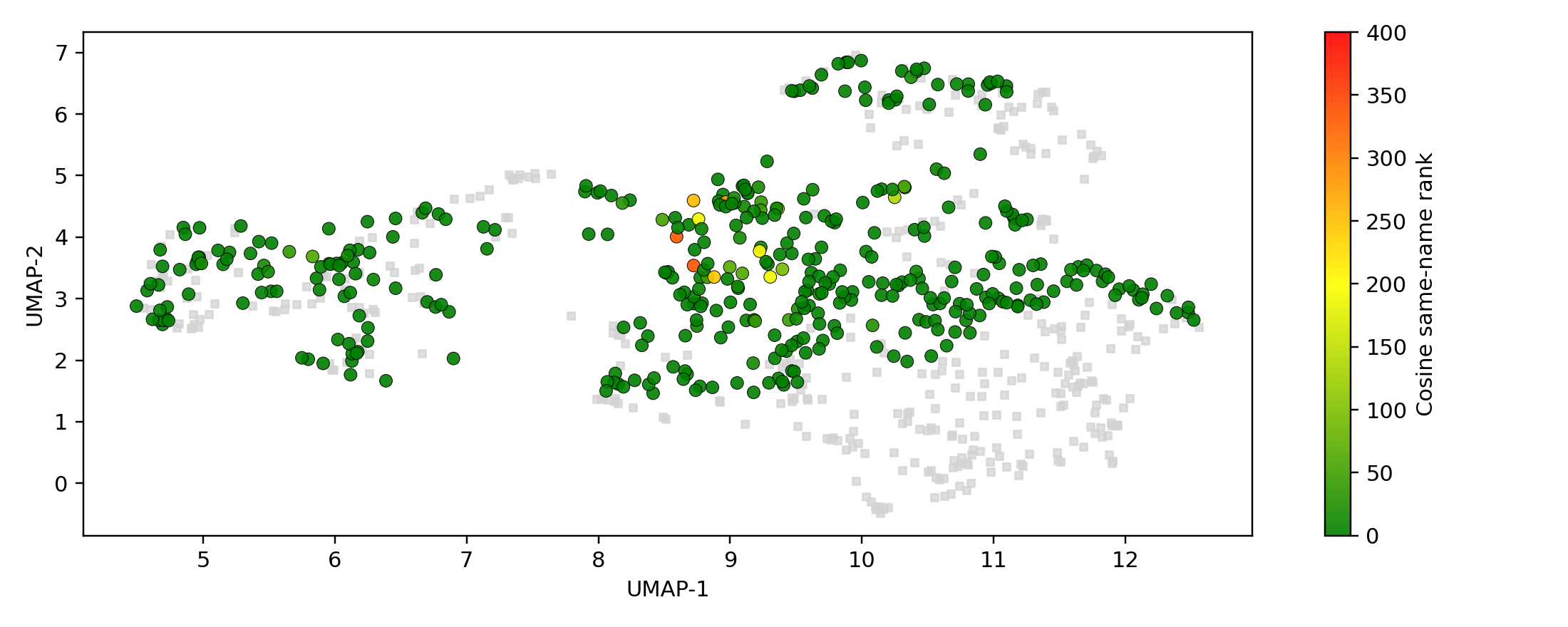}
  \caption{\textbf{UMAP projections for ARC-1.} \defaultTTT embeddings (left) are projected far from the original embedding distribution, while \ourTTT embeddings (right) are projected adequately onto it. The original embedding usually ranks among the closest train embeddings to the fine-tuned one with \ourTTT (most points are dark green, indicating a good rank).}
  \label{fig:ttt_comparison}
\end{figure*}

\Cref{fig:ttt_comparison} compares UMAP projections of \defaultTTT and original train embeddings with projections of \ourTTT and original train embeddings, showing a clear distribution shift in the former case while \ourTTT better recaptures the structure of the manifold of the original embeddings. The ground-truth embedding is generally among the closest neighbors (computed in the original embedding space) of the ``test'' \ourTTT-produced embedding under consideration (most points are dark green), while \defaultTTT embeddings are generally much further away from the original embeddings. 
To assess whether the local geometry of ARC-1 task embeddings is preserved (which could happen regardless of retrieval performance), we compared \(k\)-nearest-neighbor structure between the original train embedding space and embeddings from \ourTTT and \defaultTTT. For each task, we computed top-\(k\) neighbors in each space (excluding self), then measured neighborhood agreement with the original space via set overlap proportion. We found a large and consistent gap in favor of \ourTTT: mean overlap at $k=10$ is 33\% for \ourTTT vs 4\% for \defaultTTT (with similar observations across \(k \in \{1...50\}\)). These results indicate that \ourTTT preserves local relational structure from the pre-trained task-embedding manifold, whereas \defaultTTT distorts it.


\begin{table}[t]
\caption{\textbf{Quantitative implicit rule induction.} We evaluate \ourTTT and \defaultTTT on \ARClike datasets in their ability to produce embeddings that potentially elicit rule induction. It appears that \ourTTT excels, while \defaultTTT generally loses any semantically meaningful information.}
\label{tab:perf-results}
\centering
\small
\setlength{\tabcolsep}{4pt}
\begin{tabular}{lclcccc}
\toprule
\multirow{2}{*}{Train dataset} & Num. tasks & \multirow{2}{*}{Evaluation} & \multirow{2}{*}{Metric} & \multicolumn{3}{c}{Metric score} \\
\cmidrule(lr){5-7}
 & train/probe &  &  & Chance & \defaultTTT & \textbf{\ourTTT} \\
\midrule
ARC-1 train & 400/same & Emb. retrieval (train) & Top-5 match $\uparrow$ & 1.3\% & 6.5\% & \textbf{86.25\%} \\
ConceptARC & 160/same & Emb. retrieval (train) & Top-5 match $\uparrow$ & 3.1\% & 69.4\% & \textbf{98.8\%} \\
ConceptARC & 160/same & Lin. classif. (train) & Accuracy $\uparrow$ & 6.2\% & 21.3\% & \textbf{60.6\%} \\
Custom LUCD & 15/150 & Lin. classif. (test) & Accuracy $\uparrow$ & 6.7\% & 11.3\% & \textbf{100\%} \\
Custom Moves & 121/1210 & Lin. regress. (test) & RMSE $\downarrow$ & - & 2.94 & \textbf{2.12} \\
\bottomrule
\end{tabular}
\end{table}

\subsection{Qualitative ``out-of-distribution'' evaluation of implicit rule induction}

\textbf{Dataset.} In addition to its 400 training tasks, ARC-1 provides 400 "public evaluation" tasks (which we will refer to here as "test tasks"), that can differ from the training tasks rule-wise, making them potentially ``out-of-distribution'' (OOD). We can thus qualitatively examine the similarity between test and training tasks to measure how well an embedding-producing method works in this OOD case. This analysis is not feasible with ConceptARC due to the absence of a train-test split, and not relevant for our custom tasks since we generated only simple in-distribution tasks.  

\begin{figure*}[t]
  \centering

  \includegraphics[width=0.47\textwidth]{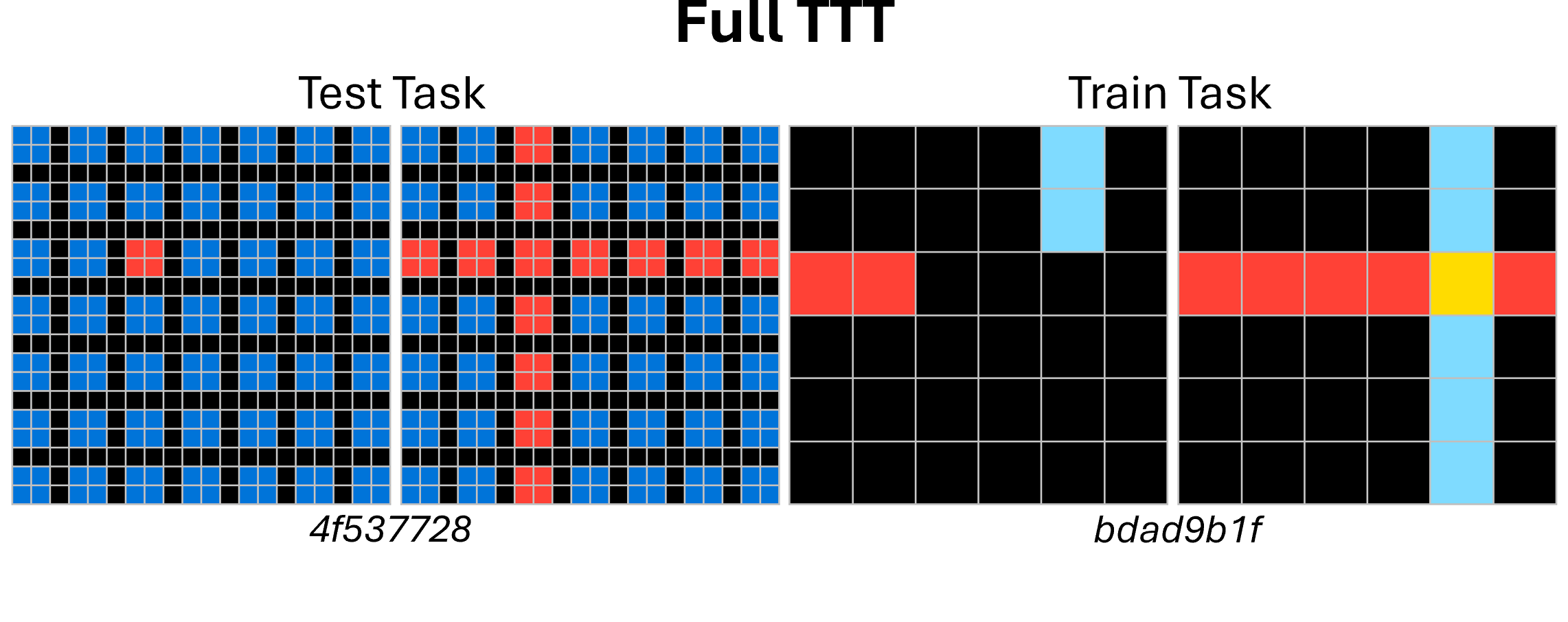}\hfill
  \includegraphics[width=0.47\textwidth]{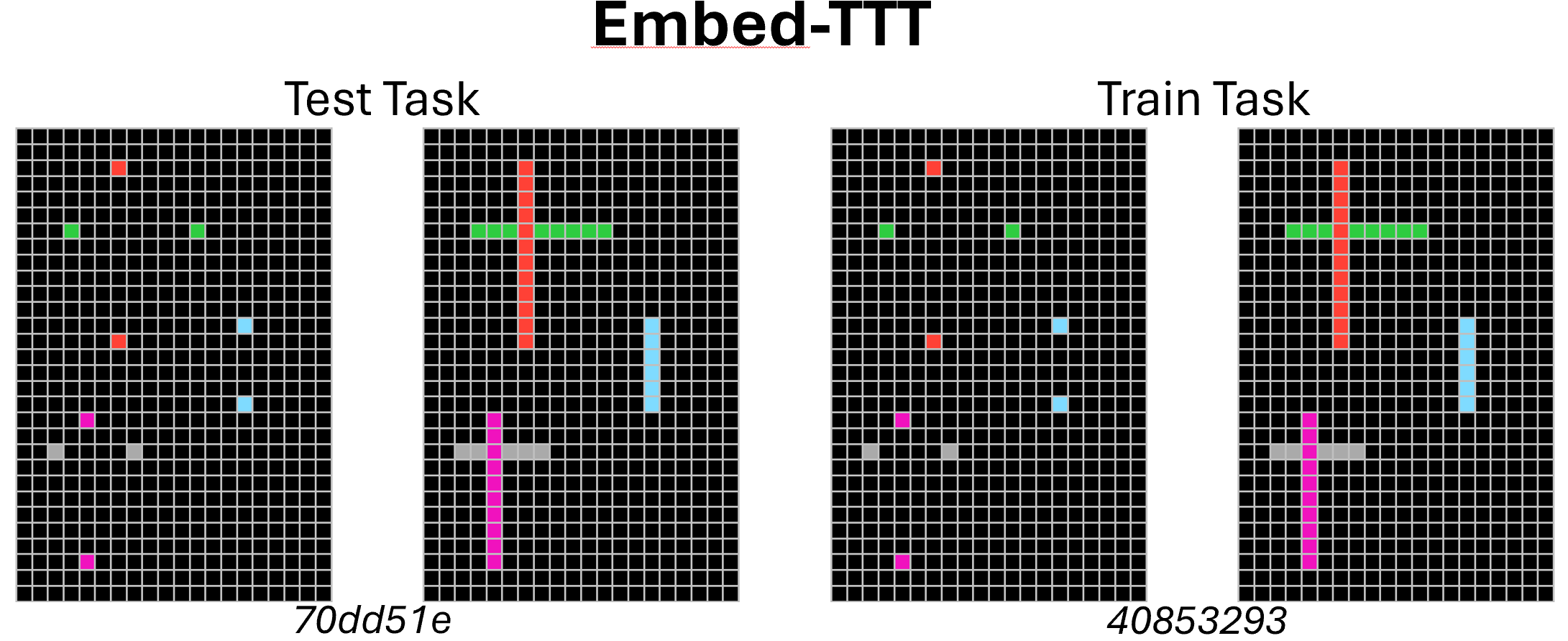}
  \includegraphics[width=0.47\textwidth]{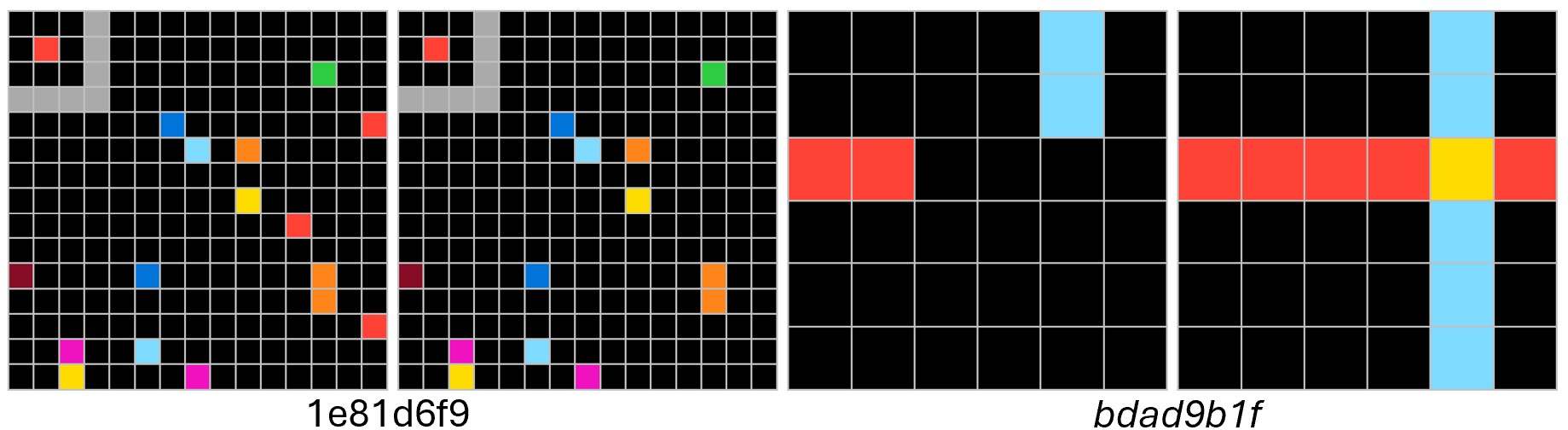}\hfill
  \includegraphics[width=0.47\textwidth]{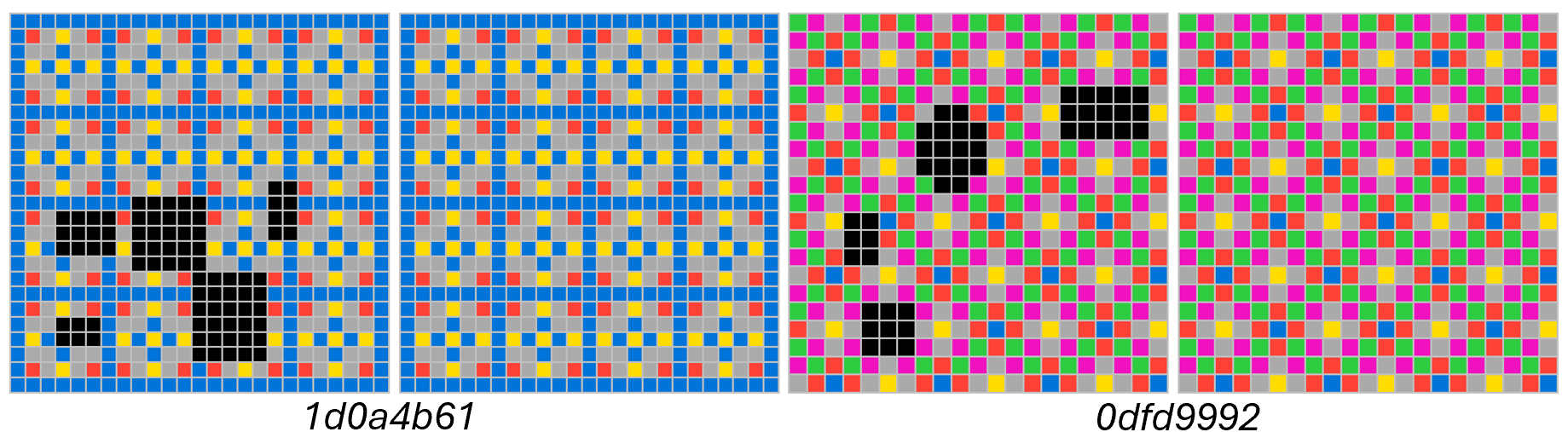}
  \includegraphics[width=0.47\textwidth]{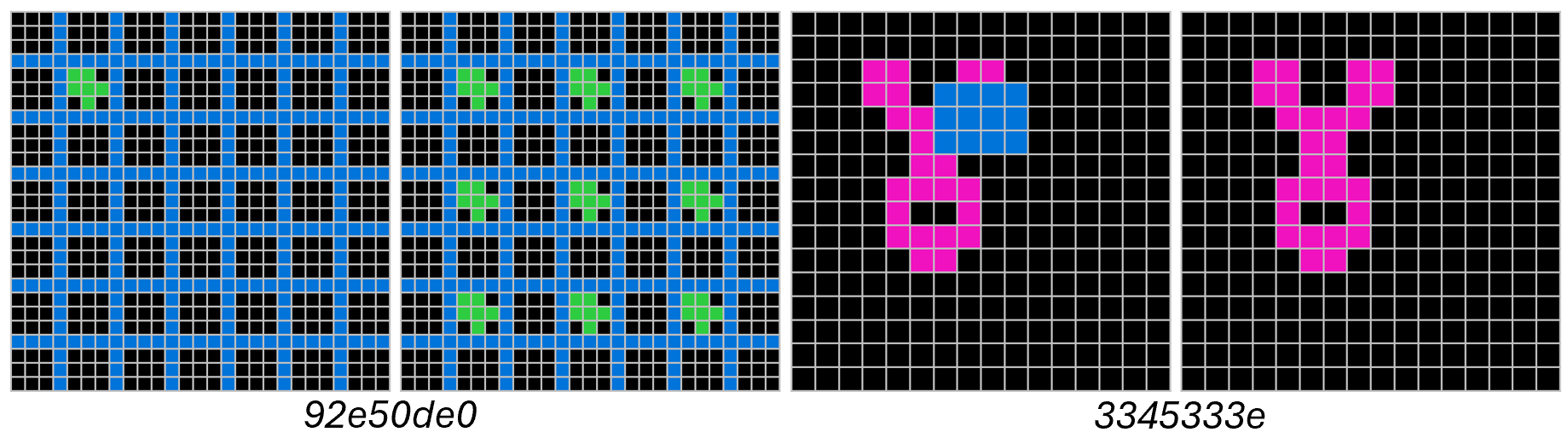}\hfill
  \includegraphics[width=0.47\textwidth]{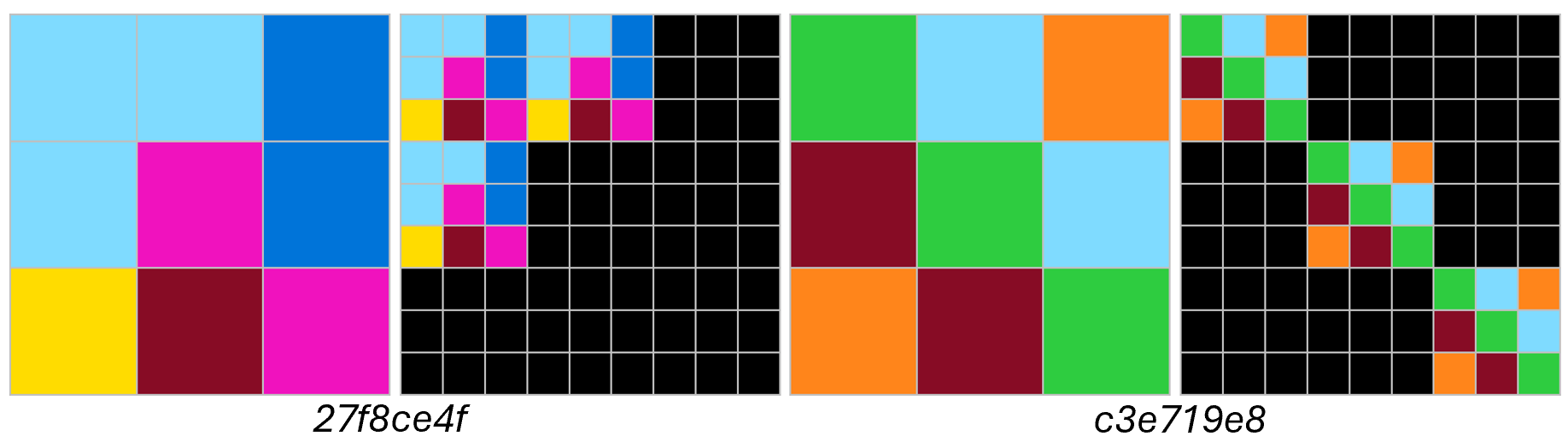}

  \caption{\textbf{Qualitative evaluation through test-train task similarity.} For \defaultTTT (left part) and \ourTTT (right part), we report the closest associations between test (left input-output pairs per part) and train (right input-output pairs) tasks. \defaultTTT does not associate tasks that seem to share common rules, while \ourTTT associates tasks that are clearly similar rule-wise, the top one being a task duplicate with different IDs (070dd51e and 40853293) in test and train sets.}
  \label{fig:task-similarity-3x2}
\end{figure*}

\textbf{Results.} For that purpose, we search for the closest train task in the embedding space for each test task. To mimic the optimization dynamic undergone by the test task in this process, the search occurs among the train embeddings obtained after \ourTTT was performed, so that embeddings obtained following the same process are compared. \Cref{fig:task-similarity-3x2} shows the closest test-train task associations from \defaultTTT (left) and \ourTTT (right). We observe that \defaultTTT does not find relevant task associations, likely because of the distribution shift between test and train embeddings is too large, as already seen in~\Cref{fig:ttt_comparison} (more are shown in~\autoref{sec:retrieval-supp}). On the other hand, \ourTTT yields test-train associations that are relevant, which is non-trivial despite the good in-distribution results. In particular, the first association is an exact task duplicate shared between the test and the train sets of ARC-1. The other tasks shown clearly display the same rule semantics, showing a regular inpainting task and a copy task according to the dominant color of the input grid. In our observations, close associations are usually relevant, while farther ones are less so, as expected given that some ARC-1 test tasks differ radically from the training tasks. We did not quantify this observation further, as it is relatively subjective. Still, qualitative examination shows that \ourTTT enables meaningful comparison between test and train tasks, which is not the case with the default \defaultTTT approach.


\textbf{Test performance predictability.} We found that test tasks very close to train tasks (top 10\% closest) were often solved (resp. failed) when their nearest train neighbors were solved (resp. failed), while very distant test tasks were often failed. However, this pattern did not generalize to most tasks: distance to the nearest train task alone was usually not a reliable predictor of success. We also tested standard alternatives as success predictors (KNN, medoids, classifiers, etc.) and found no strong predictive signal. A likely explanation is that many test tasks are out-of-distribution, though this cannot be confirmed without explicit test-rule annotations.

\subsection{Cross-dataset capabilities}

\textbf{Results.} By acting as an implicit rule inductor, \ourTTT can be used to test whether a model trained on an \ARClike dataset can induce and execute rules underlying tasks of another \ARClike dataset. In that regard, \Cref{tab:passk} compares the performances of the original VARC ViT model with \defaultTTT, \ourTTT, and \embedfullTTT on ARC-1, ConceptARC and Mini-ARC~\citep{kim2022playgrounds} tasks. We report the proportion of test tasks solved at pass 1 and pass 2 in the test datasets, as per the default ARC-1 evaluation protocol. 

We observe that \embedfullTTT---finetuning the task embedding first, freezing it, then finetuning the model backbone---provides a moderate performance improvement on ARC-1 test set over the default \defaultTTT and a larger improvement on ConceptARC and Mini-ARC. These improvements might be due to a better conditioning on the backbone finetuning based on an improved embedding. Interestingly, \ourTTT, by finetuning only the task embeddings at test time, which accounts for $<0.01\%$ of the model weights, suffices to score more than 13\%  on ARC-1 at pass 1 (resp. 17\% pass 2), 8\% (resp. 14\%) on ConceptARC, and 22\% (resp. 34\%) on Mini-ARC. This indicates that the VARC model has learned general rule induction and execution capabilities, but also that some ARC-1 test tasks, ConceptARC and Mini-ARC tasks might be similar to the ARC-1 training tasks. These percentages also represent a significant fraction of what can be achieved by the model when the backbone finetuning is allowed. From an efficiency perspective measured as weights to finetune, \ourTTT results are exceptionally high and indicate, to some extent, the generalizability of our approach beyond just ARC-1 tasks. As a point of comparison, we tested two recent open-weight language models on the same public ARC-1 test set with the same pass@1 evaluation as used in our study: Llama-4-Maverick scores only 4.2\% and Kimi K2.5 scores 27.8\%. These results show that our approach outperforms these models on the same controlled settings. Proprietary models used in the official ARC leaderboard are evaluated on a private dataset and thus might have been trained on the test set considered here, hence we refrain from running expensive and potentially flawed comparisons, while acknowledging that such models are probably substantially more capable.  

Let us note that~\Cref{tab:passk} also shows that, for ConceptARC, independent performance across test input-output pairs (480) is significantly higher than when aggregated per task (3 test pairs per task). This implies that the model does not solve ConceptARC tasks on a consistent basis, with successful test samples scattered across the tasks rather than consistently succeeding or failing on specific tasks. For ARC, a similar analysis gives a boost of only $0.3\%$ across all metrics, mainly because most test tasks contain only one test sample, making the evaluation almost equivalent in both cases. 

Interestingly, even though VARC is a highly capable model trained on many ARC tasks and extra RE-ARC~\citep{Hodel2024addressing-arxiv} samples, it does not necessarily learn the simple transformations involved in LUCD or Moves. Indeed, with \ourTTT on LUCD, pass@1 is 56\% and pass@2 is 67.3\%, on Moves, pass@1 is 11.6\% and pass@2 is 19.8\% (and of course any full finetuning reaches 100\% but the expectation was to reach that with just task embedding finetuning). This indicates that models trained on some ARC-like tasks primarily ``interpolate'' between the contained functionalities, while even simple OOD tasks might not be directly solved by just finetuning the task embedding. This point is further reinforced in~\cref{sec:MovesStructure}. This is a curious property, and points to how brittle models might be even when they appear strong on benchmarks claiming to require OOD generalization.

Finally, additional experiments with \ourTTT on ARC-1 test set showed that performance could be pushed slightly further. By increasing the ``breadth'' of search by finetuning up to 8 different task embeddings at test time (instead of just one) and then using the one with the lowest training loss, pass@1 (resp. pass@2) reaches 14.8\% (resp. 18.5\%). Further increasing the ``depth'' of search by finetuning two task embeddings applied sequentially allows to reach up to 17.3\% (resp. 20.3\%). Such results show that the trained model still contains further capabilities, yet the gains of continuing to further increase breadth and depth are likely marginal and would be computationally expensive. Hence, we did not investigate this direction deeper.

\begin{table}[t]
\caption{\textbf{Cross-dataset results.} Our \embedfullTTT allows to reach better final per-task performances than \defaultTTT on ARC-1, ConceptARC and Mini-ARC. In addition, \ourTTT, solves a significant amount of tasks by just finetuning the embeddings, indicating that rules underlying test tasks might not always differ much from train tasks. Pair-level (instead of task-level) evaluation on ConceptARC(*) shows a large gap, indicating potential difficulties in solving all input-output pairs of a given task consistently. Finally, ConceptARC concepts inferred on ARC-1 tasks by task embeddings similarity decently match manual annotations despite the fact that the model was never trained for concept retrieval.}
\label{tab:passk}
\centering
\small
\begin{tabular}{llcccccc}
\toprule
\multirow{2}{*}{Train dataset} & \multirow{2}{*}{Test dataset} & \multicolumn{3}{c}{Pass@1 per TTT method} & \multicolumn{3}{c}{Pass@2 per TTT method} \\
\cmidrule(lr){3-5}\cmidrule(lr){6-8}
 &  & Full & Embed & Embed-Full & Full & Embed & Embed-Full \\
\midrule
ARC-1 train & ARC-1 test & 48.8\% & 13.0\% & \textbf{49.5\%} & 53.8\% & 17.2\% & \textbf{55.2\%} \\
ARC-1 train & ConceptARC & 23.8\% & 8.8\% & \textbf{28.1\%} & 32.5\% & 14.4\% & \textbf{37.5\%} \\
ARC-1 train & Mini-ARC & 59.7\% & 22.2\% & \textbf{63.8\%} & 62.4\% & 34.9\% & \textbf{71.1\%} \\
\noalign{\vskip 2pt}
\hdashline
\noalign{\vskip 2pt}
ARC-1 train & ConceptARC (*) & 47.1\% & 24.0\% & \textbf{49.8\%} & 53.8\% & 33.3\% & \textbf{57.1\%} \\
\midrule
\multicolumn{2}{l}{Concept retrieval in ARC} & \multicolumn{6}{l}{Top-1: Test 36.3\% (29/80), Train 35.0\% (28/80), from \ourTTT} \\
\bottomrule
\end{tabular}
\end{table}

\textbf{Concept retrieval in ARC.} In addition, given that they were obtained by using the same backbone, ARC-1 and ConceptARC embeddings lie in comparable regions of the embedding space (akin to \Cref{fig:ttt_comparison} (right)). Hence, we can compare them, by assigning each ARC-1 task its closest ConceptARC task, using \ourTTT-produced embeddings for ConceptARC tasks, ARC-1 train and ARC-1 test tasks. In particular, we want to know if the core concepts close to ARC-1 tasks are indeed related to the tasks. For that purpose, we manually annotated a balanced subset of 80 ARC-1 train tasks and 80 ARC-1 test tasks (5 for each of the 16 concepts uniformly spread across cosine distances), by indicating which concepts (at most 3) were best related to the tasks. We then check if the closest concept found matches one of the annotations. The results are reported in~\Cref{tab:passk} and show decent agreement: $36.3\%$ on test and $35.0\%$ on train ARC-1 tasks. Importantly, the agreement is stronger for closer embedding associations: in both train and test splits, the closest-distance quartile yields the highest agreement ($60\%$ on test, $50\%$ on train). Overall, nearest ConceptARC neighbors provide a useful interpretability signal, especially at small distances, but do not suffice as concept classifiers for ARC-1 tasks. However, we noticed that many ARC-1 tasks are not perfectly aligned with the predefined concepts, and that additional concepts would improve coverage. Still, the results are far above pure chance (about $6.3\%$). Hence, \ourTTT enables, for the first time in this setting, a practical examination of meaningful connections between different \ARClike datasets. 

\subsection{Compositional Structure of the Embeddings}
\label{sec:MovesStructure}

\textbf{Datasets and scope.} Using our Moves dataset, we further examine how the model structures the embedding space. For that purpose, we consider the cosine similarity of composed transforms (e.g. R3-U2) to their underlying atomic transforms (R3-U0, R0-U2). Then, we investigate which tasks are solvable with \ourTTT from a subset of the initial training dataset.

\begin{figure*}[t]
  \centering
  \newlength{\movefigH}
  \setlength{\movefigH}{0.14\textheight}
  \includegraphics[height=\movefigH,keepaspectratio]{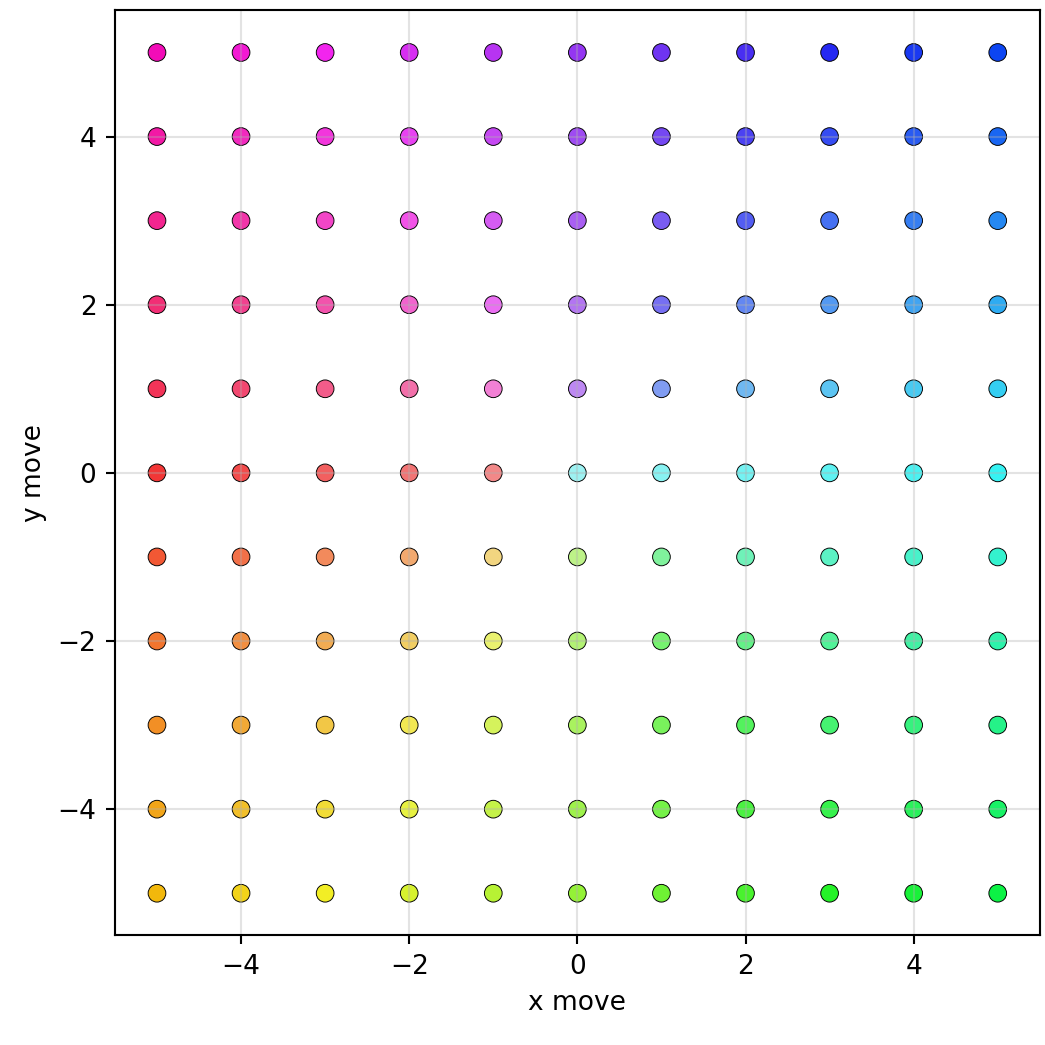}
  \includegraphics[height=\movefigH,keepaspectratio]{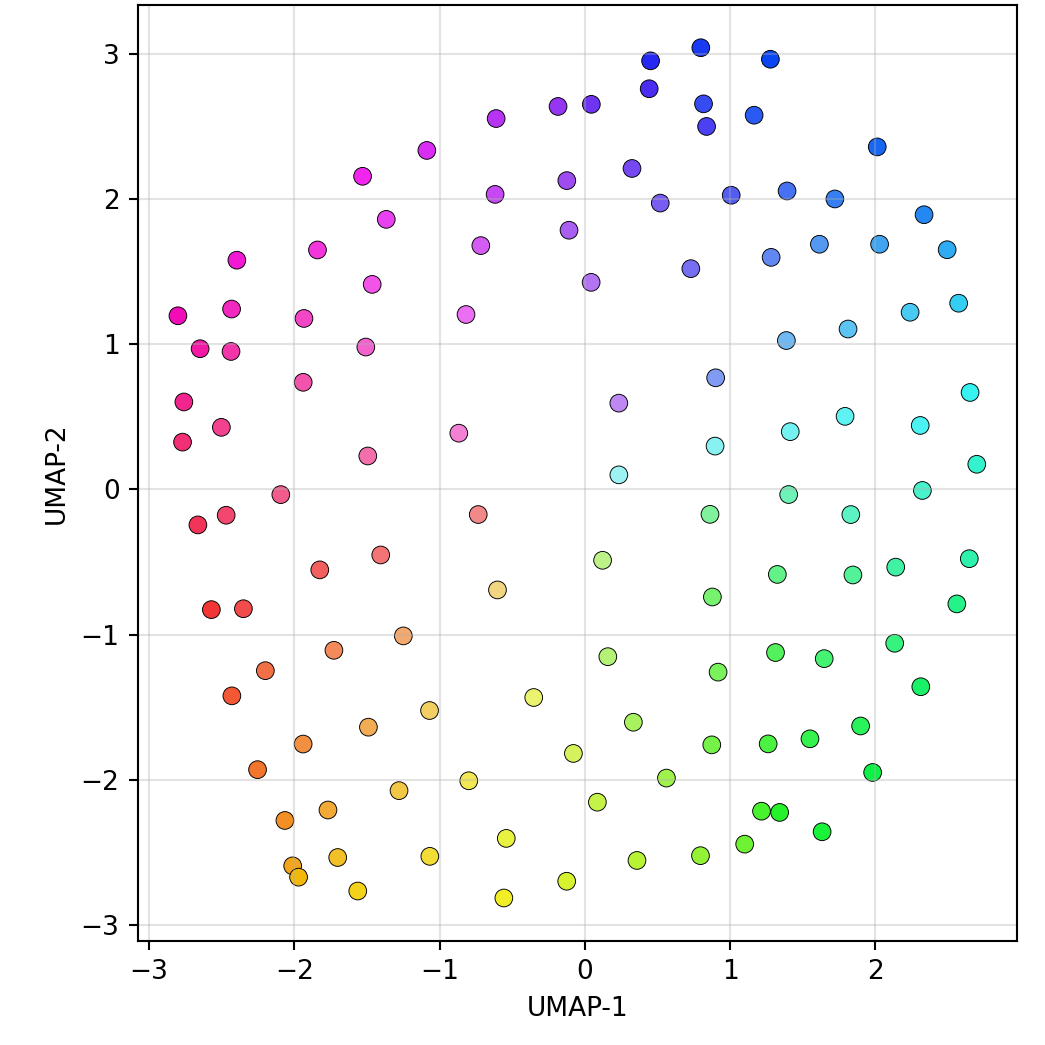}\hfill
  \includegraphics[height=\movefigH,keepaspectratio]{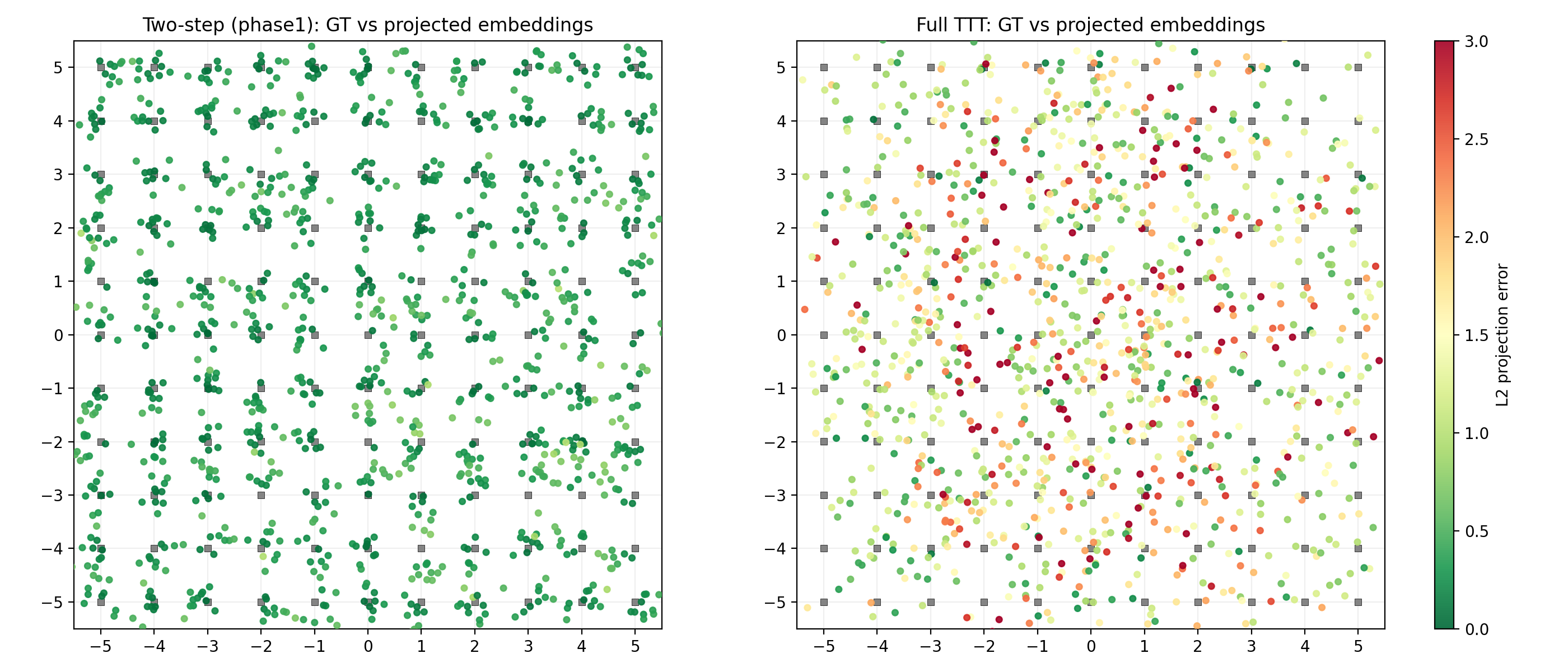}
  \caption{\textbf{Embeddings structure.} In our Moves dataset, the task geometry (left) is reflected by the UMAP of train embeddings (center-left), which forms an almost regular grid. We then learn a linear map from train embeddings finetuned with \ourTTT (resp. \defaultTTT) to their ground-truth move coordinates (gray squares) (center-right, resp. right), and apply it to test embeddings (points colored by projection error). \ourTTT preserves a regular grid with low error, unlike \defaultTTT.}

  \label{fig:move-tasks-1x3}
\end{figure*}

\textbf{Compositions vs. atoms.} Across the 100 composed transforms, we find that the two atoms are both the most or second-most similar purely horizontal and vertical atoms for 94\% of the cases. The cosine similarities with ``unit'' transforms (in the example case R1-U0 and R0-U1) are much weaker, indicating a lack of scaling capability, in favor of a plain recombination of individual ``already-scaled'' atomic projections. The parametric nature of the dataset enables an intuitive visualization:~\Cref{fig:move-tasks-1x3} (left and center-left) clearly shows that the UMAP projection of the \ourTTT embeddings follows an almost regular grid-like pattern that mimics the original geometry of the dataset, where each task of the dataset can be associated to its move coordinates in the plane. Similar visualizations from PCA are provided in~\autoref{sec:pca-moves}. Further, a linear mapping learned from \ourTTT train embeddings efficiently projects test embeddings onto the grid, contrary to a mapping learned from \defaultTTT train embeddings (right-center and right panels).

\begin{figure*}[t]
  \centering
  \newlength{\mtwoH}
  \setlength{\mtwoH}{0.2\textheight} 

  \includegraphics[height=\mtwoH,keepaspectratio]{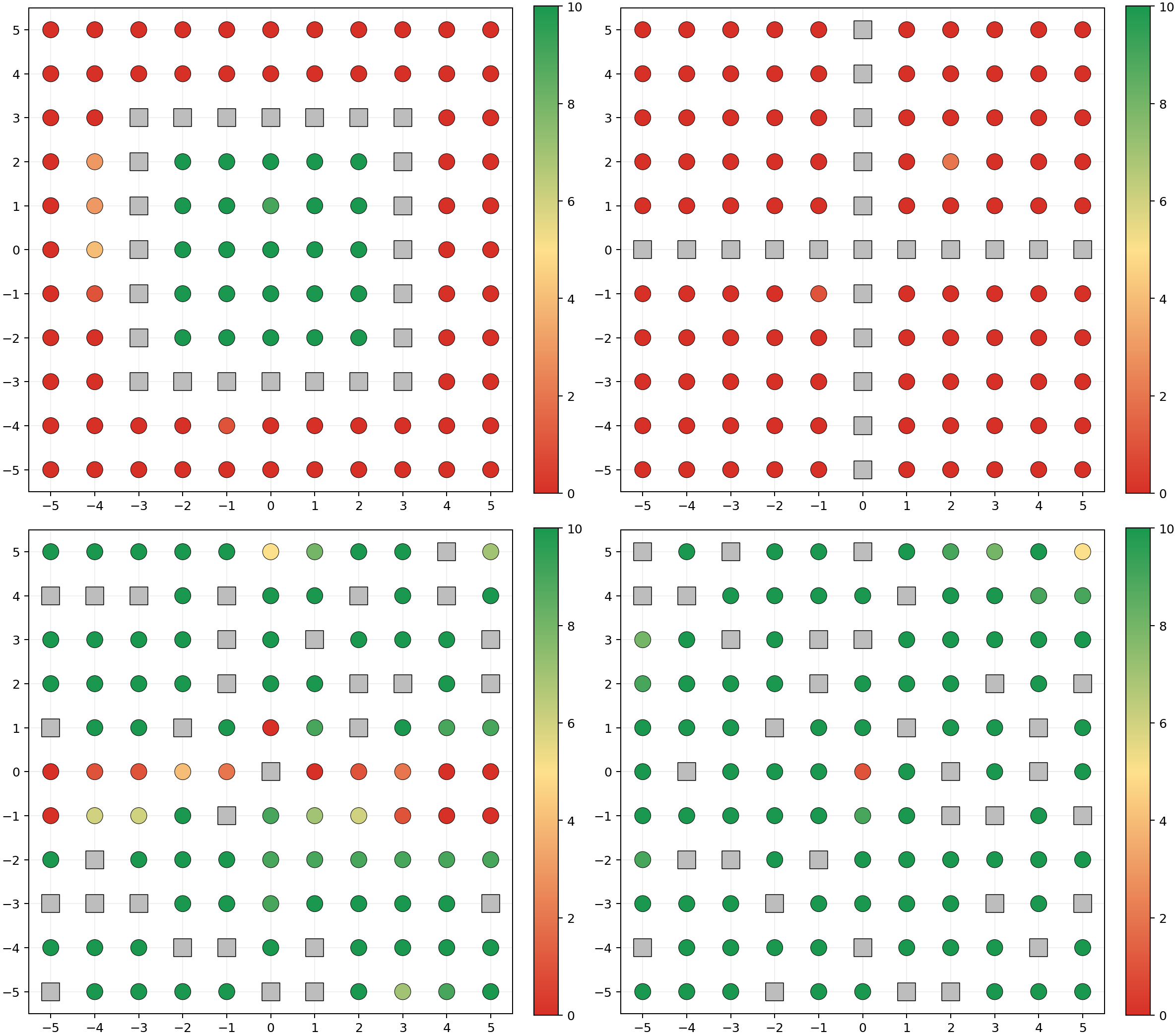}\hfill
  \includegraphics[height=\mtwoH,keepaspectratio]{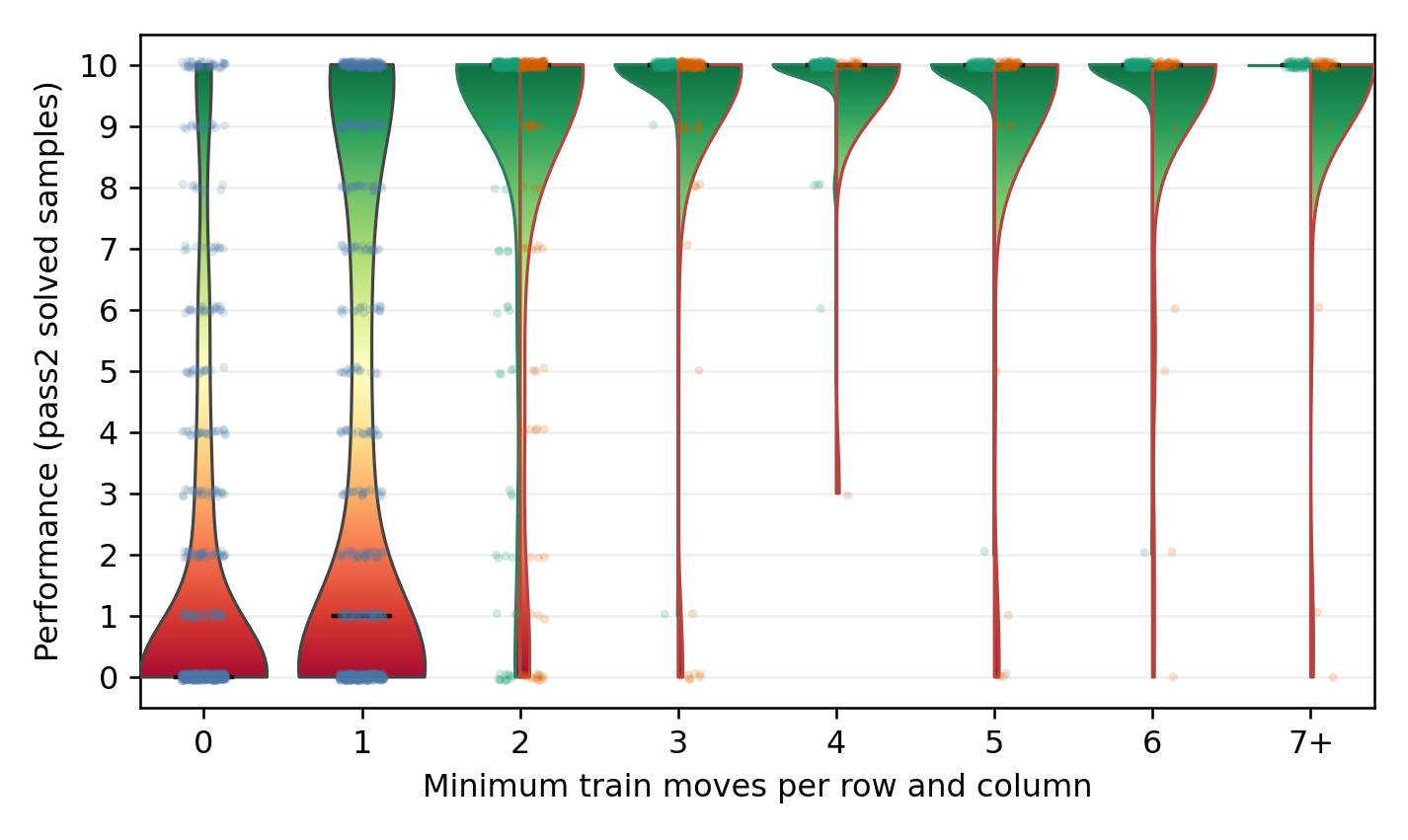}

  \caption{\textbf{Task inter/extrapolation.} (Left) Move tasks are placed in the plane by their displacement. Gray squares denote train tasks; circles denote test tasks, colored by success under \ourTTT finetuning, across several train/test splits. Test tasks are mostly solved when at least $2$ train moves lie on both their row and column, suggesting interpolation is feasible while extrapolation is limited. (Right) Violin plots aggregate test successes by the minimum number of train tasks on the same row/column. A clear transition appears at $2$, akin to a ``grokking'' effect. Split violins show ``inside'' test tasks (left half) are slightly better solved than ``outside'' tasks (right half).}
  \label{fig:move-tasks2-side-by-side}
\end{figure*}

\textbf{\ourTTT allows interpolation between tasks.} For the Custom Moves dataset, we further investigate which moves might be sufficient as training set so that the model generalizes to the remaining unseen moves. In that regard,~\Cref{fig:move-tasks2-side-by-side} (left) shows various configurations, where training moves are represented in gray squares while test moves are circles colored according to the model performance on these moves with \ourTTT. The top left panel indicates that interpolation inside known moves is achievable at test time, while extrapolation outside known moves is not possible in that configuration. The top right panel shows, however, that, even if the model was trained on all move types in each direction, it is not able to compose them without having seen any composition beforehand. The bottom left panel shows 30 randomly selected train moves and indicates that test moves located on lines where only one training move was present are less easily achievable. Finally, the bottom right panel shows that, when 3 training moves are present on each row and each column (33 train moves in total), almost all the other moves can be solved at test time by \ourTTT, at the exception in this case of the task involving no move, likely because the model was trained to always do at least one directional transform. With comparable number of train moves (30 vs 33), this hints that train moves configuration can play a major role into generalization capabilities.  

These observations suggest that the minimal number of training moves present in the same rows and columns as the test moves is a strong predictor of success on test moves. Concretely, for each test move (across all our experiments, including runs not shown), we compute the minimum value between the number of training tasks on its row and on its column, then group test performances by this value (from 0 to 10, shown in violin plots in~\Cref{fig:move-tasks2-side-by-side}, right).  A clear transition appears at $2$: the model tends to ``grok'' once at least $2$ aligned training moves are available. A plausible explanation is that reusing embedding dimensions for a given direction requires repeated evidence; with only one instance, there is little pressure to factorize shared structure. The violin plots are further split into two parts vertically. The left-hand side corresponds to test moves that are ``inside'' the (at least) two training moves located on the same row or column, the right-hand side for ``outside'' test moves. It appears that test moves located directly inside train moves are more easily solved, while those outside are occasionally failed. Let us note that, even when they are solved, they do not correspond to extrapolation capabilities, since they still lie within the convex hull of the training moves and simply need to reuse known dimensions of the embeddings to encode the correct transform. Further, we find that, when adding a third transformation that introduces simple color change, even when no compositions are shown, \ourTTT enables the correct projection to a point that lies closest to both valid moves and color changes, whereas \defaultTTT fails completely (see \autoref{sec:move_switch}).

We note a fundamental limitation in that \ourTTT does not seem capable to add moves together. For instance, in the top left panel of \Cref{fig:move-tasks2-side-by-side}, the model learned to move by 1 unit to the right (through the train moves located at (+1,+3) and (+1,-3)) and obviously by 3 units to the right. However, we found no way (heuristically or by constrained optimization) to create an embedding that activates both +3 and +1 moves to the right to achieve +4. The issue likely lies in the execution of the embedding itself, because the backbone was never trained on such combinations. We also found that forcing the model to apply two moves iteratively that could be optimized jointly was unsuccessful. It is likely that the lack of prior training on such iterative composition causes this failure. 

\section{Conclusion}

We introduced \ourTTT, a test-time training protocol for Vision ARC (VARC) that first optimizes task embeddings to solve \ARClike tasks, before an optional backbone finetuning. Across ARC-1, ConceptARC, and controlled datasets with known rules, \ourTTT yields task embeddings that are more semantically meaningful and substantially more rule-aligned than those from the original full finetuning protocol. We show that these embeddings enable test-train task associations, cross-dataset task comparisons, and task geometry recovery. As limitations, \ourTTT requires iterative test-time optimization through the full model, and the two-step process is slower to reach only slightly better performances. However, the value of \ourTTT embeddings resides mainly in their rule induction capabilities, supported by our results and to be distinguished from rule execution by the backbone.

Our findings also convey a broader message for evaluating \ARClike benchmarks: final accuracy alone does not suffice to characterize reasoning. More focus on assessing rule induction is needed, with a clear separation between in-distribution and out-of-distribution at the rule level, as rigorously done in common machine learning practice. Eventually this will lead to more reliable claims about model ``reasoning'' capabilities. 


\newpage
\begin{ack}
A. Deliège is a Postdoctoral Researcher of the Fonds de la Recherche Scientifique – FNRS. Sandia National Laboratories is a multimission laboratory managed and operated by National Technology and Engineering Solutions of Sandia, LLC, a wholly owned subsidiary of Honeywell International, Inc., for the U.S. Department of Energy's National Nuclear Security Administration under contract DE-NA-0003525. C. Beger was supported in part through the BANYAN Institute, funded by Sandia National Laboratories' Laboratory Directed Research and Development program. The authors thank the Santa Fe Institute for hosting a visit during which part of this work was discussed.
\end{ack}


\bibliographystyle{plainnat}
\bibliography{bib/abbreviation,bib/bibliography}

\clearpage
\appendix

\section{Our Custom LUCD dataset}
\label{sec:LUCD}

\begin{figure*}[t]
  \centering
  \includegraphics[width=0.32\textwidth]{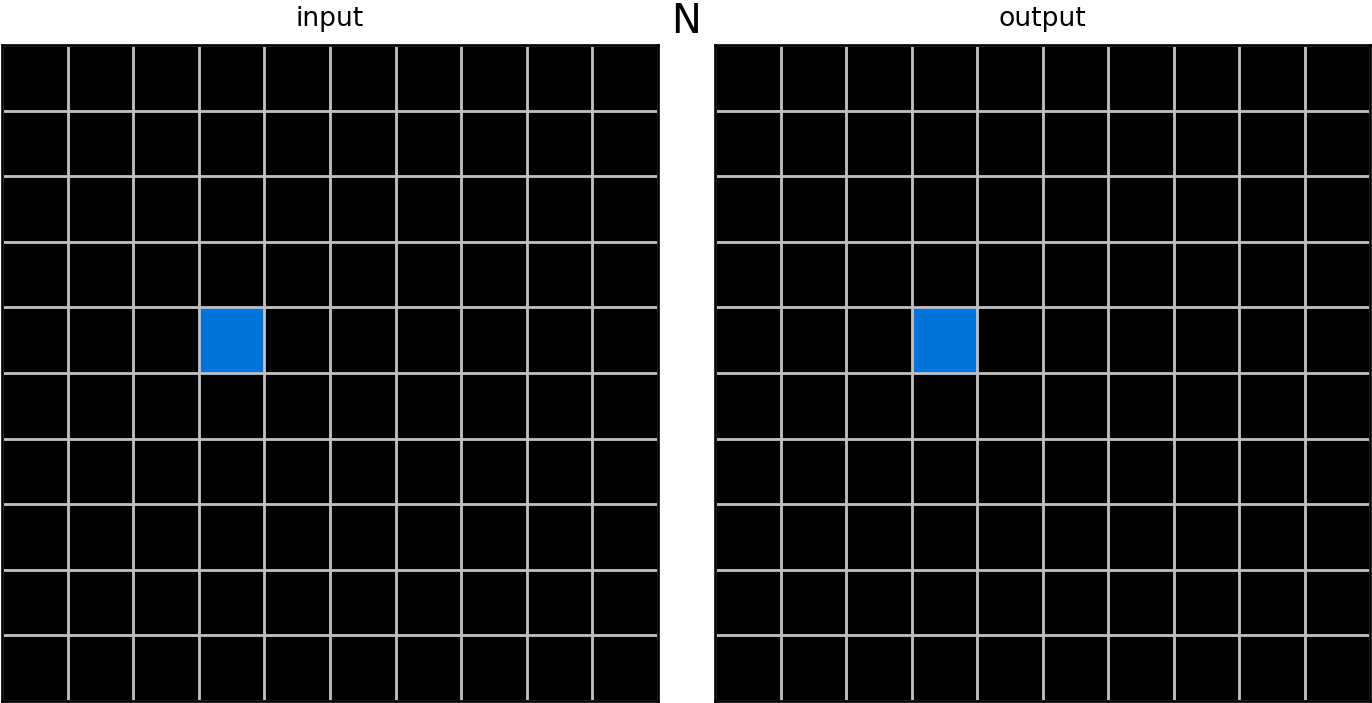}\hfill
  \includegraphics[width=0.32\textwidth]{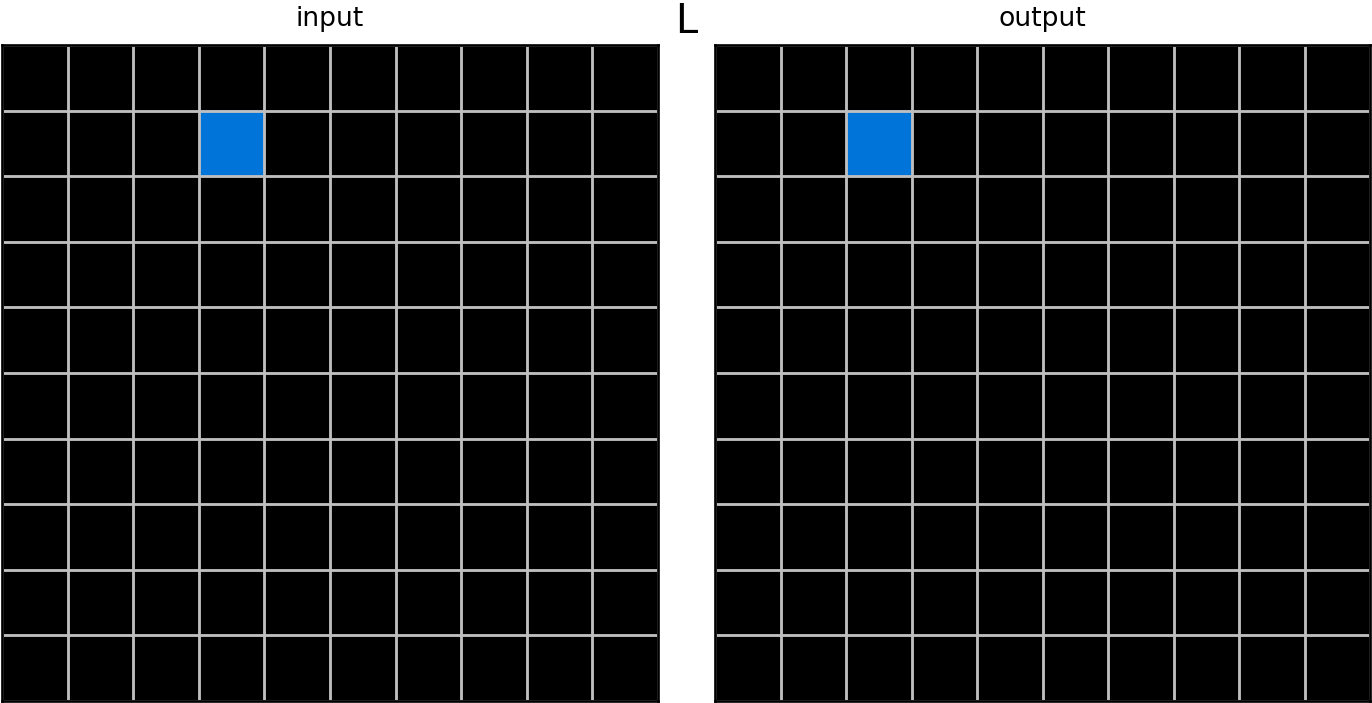}\hfill
  \includegraphics[width=0.32\textwidth]{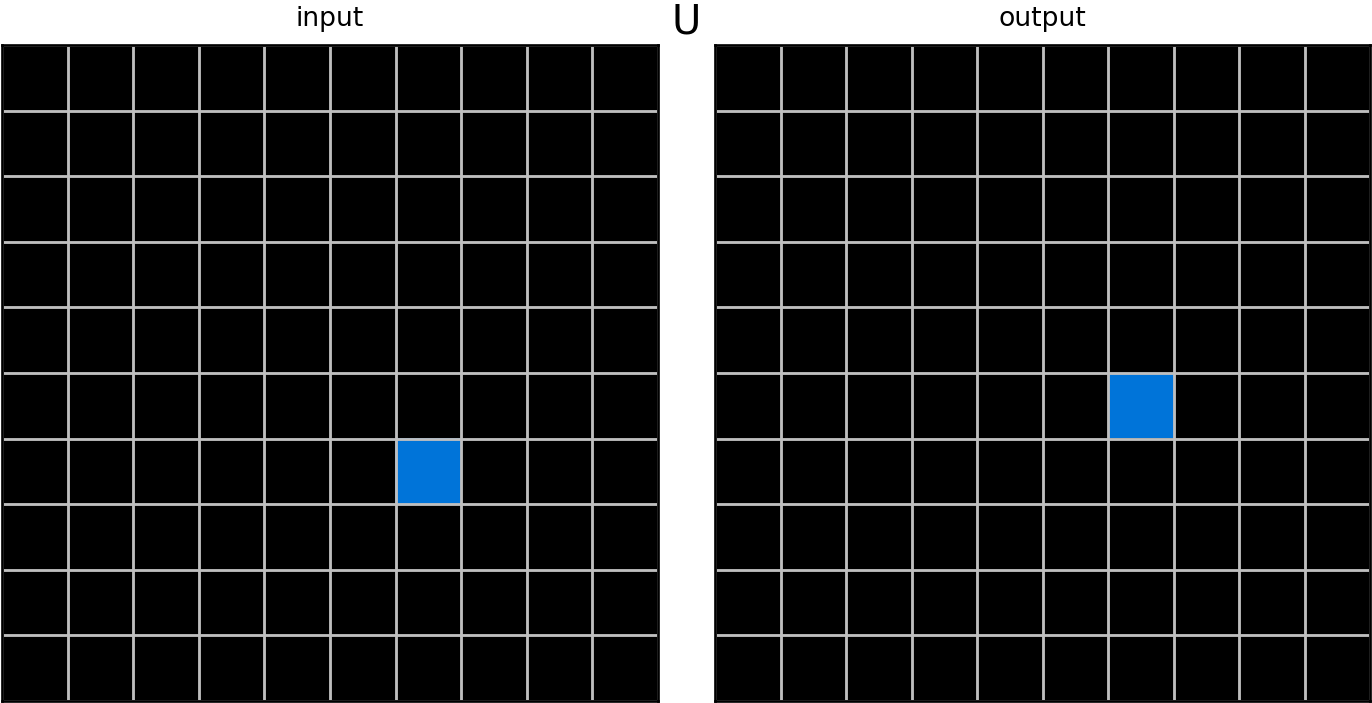}

  \vspace{0.35em}

  \includegraphics[width=0.32\textwidth]{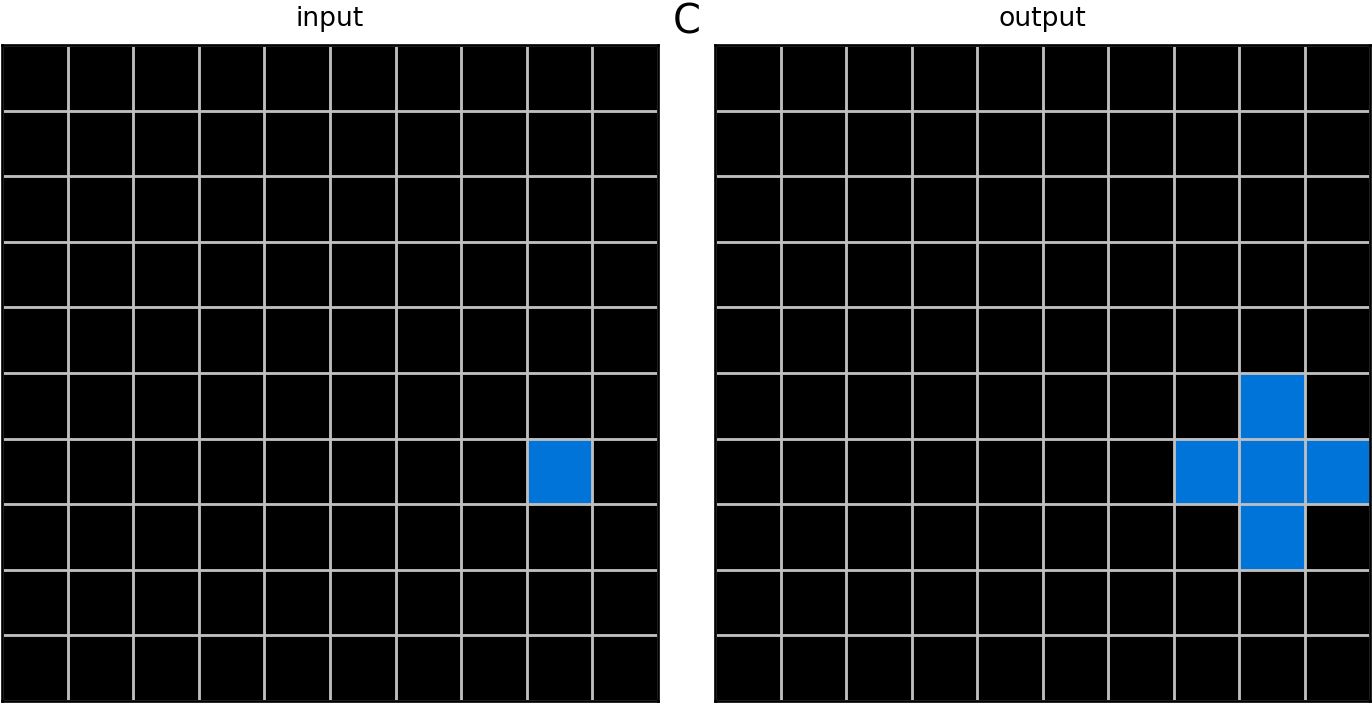}\hfill
  \includegraphics[width=0.32\textwidth]{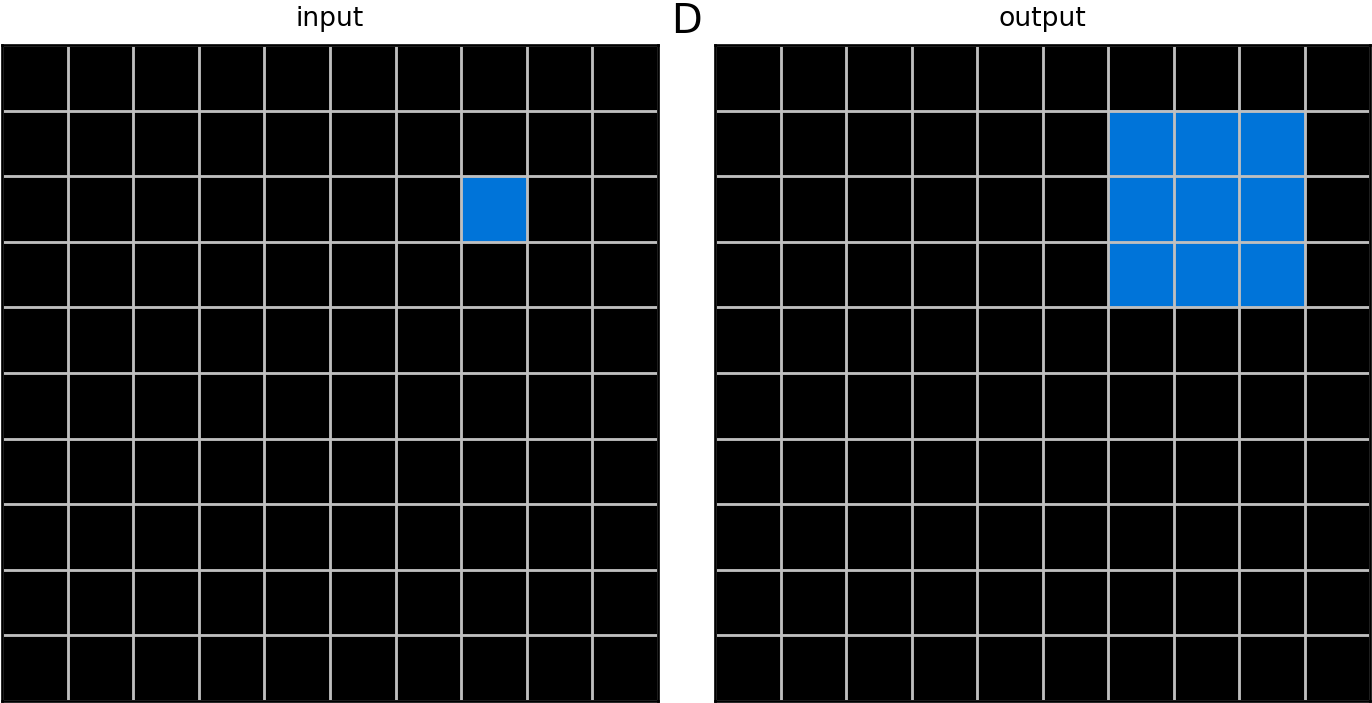}\hfill
  \includegraphics[width=0.32\textwidth]{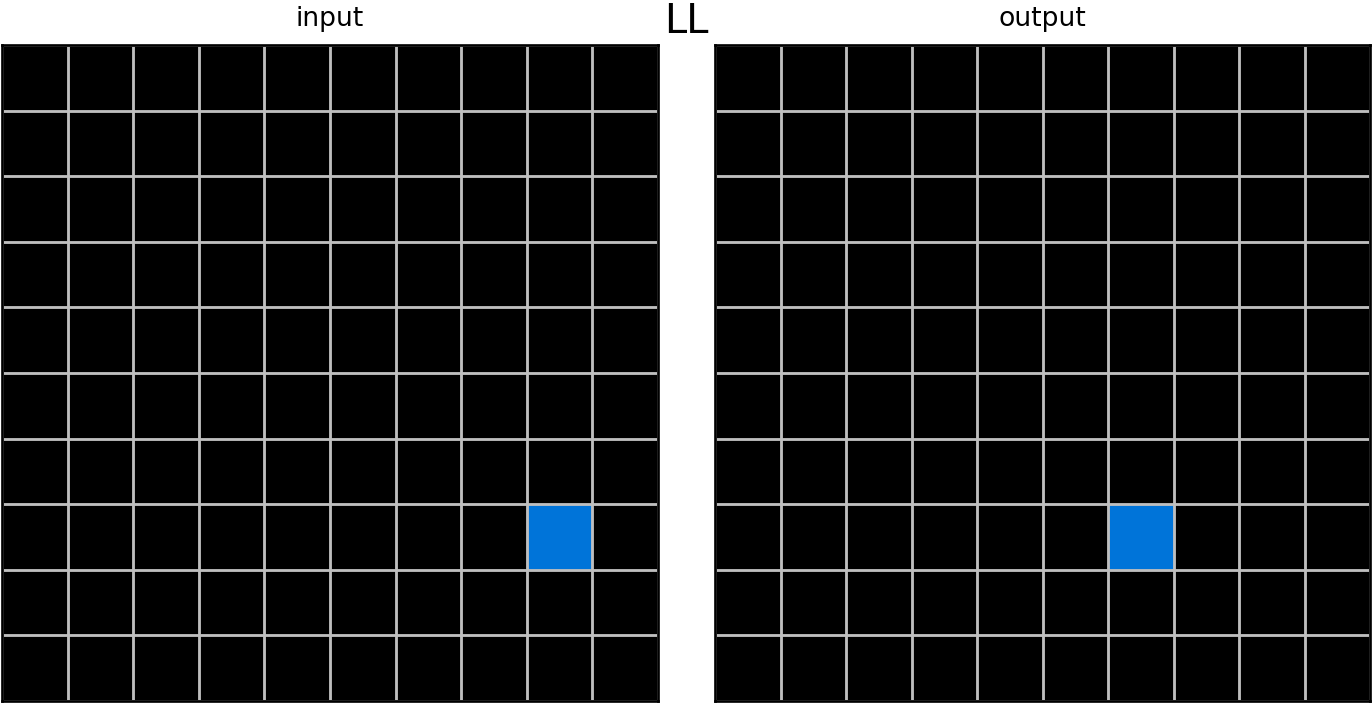}

  \vspace{0.35em}

  \includegraphics[width=0.32\textwidth]{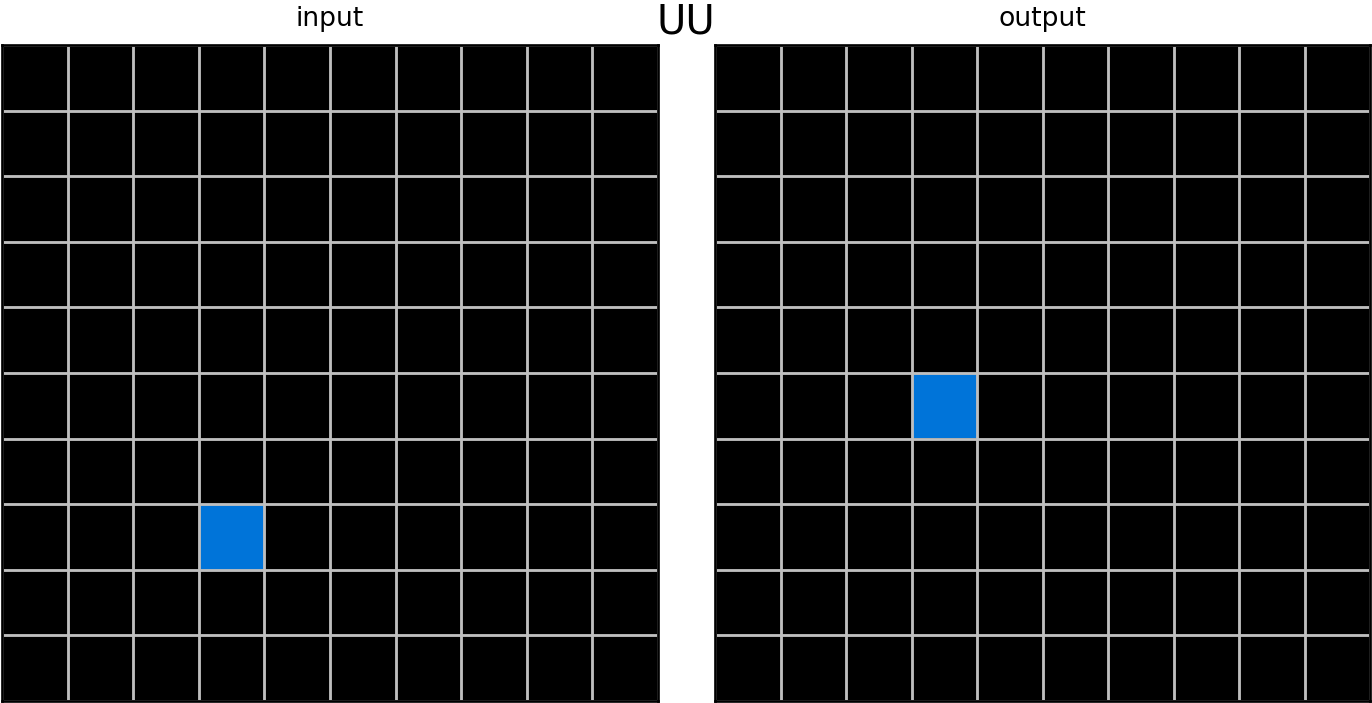}\hfill
  \includegraphics[width=0.32\textwidth]{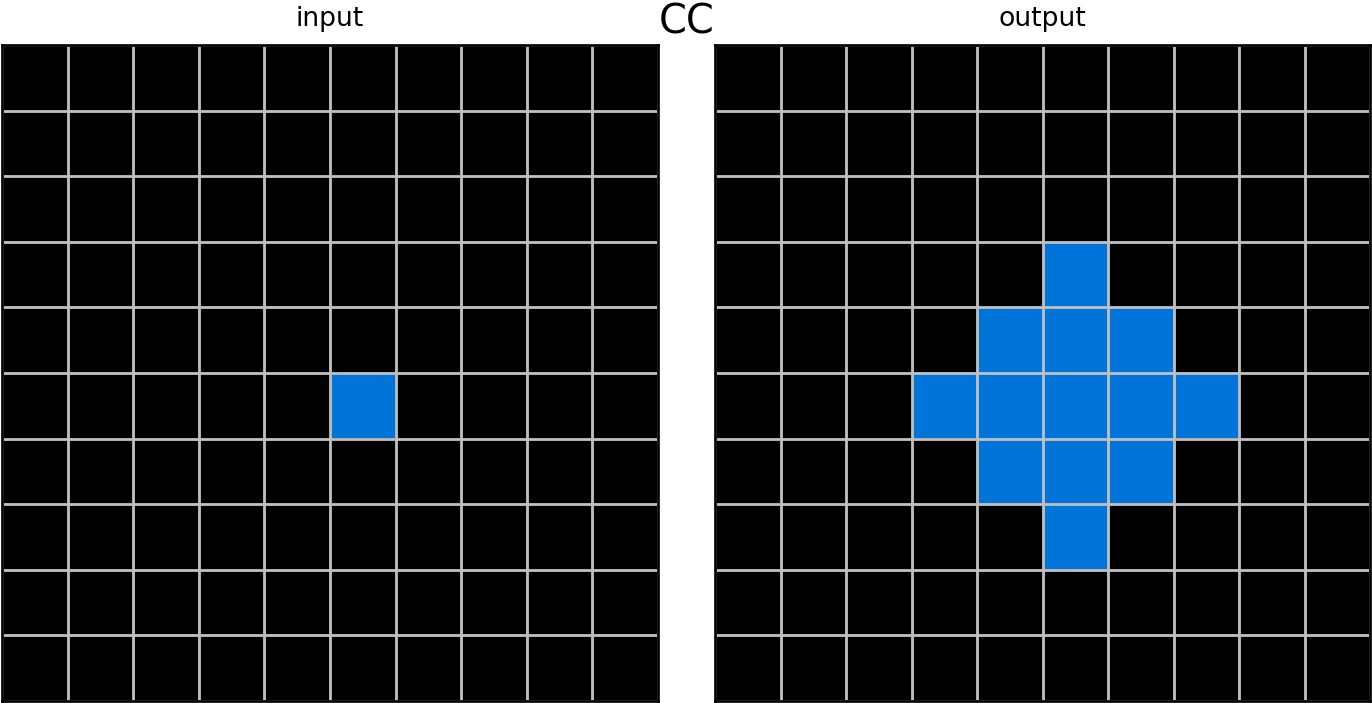}\hfill
  \includegraphics[width=0.32\textwidth]{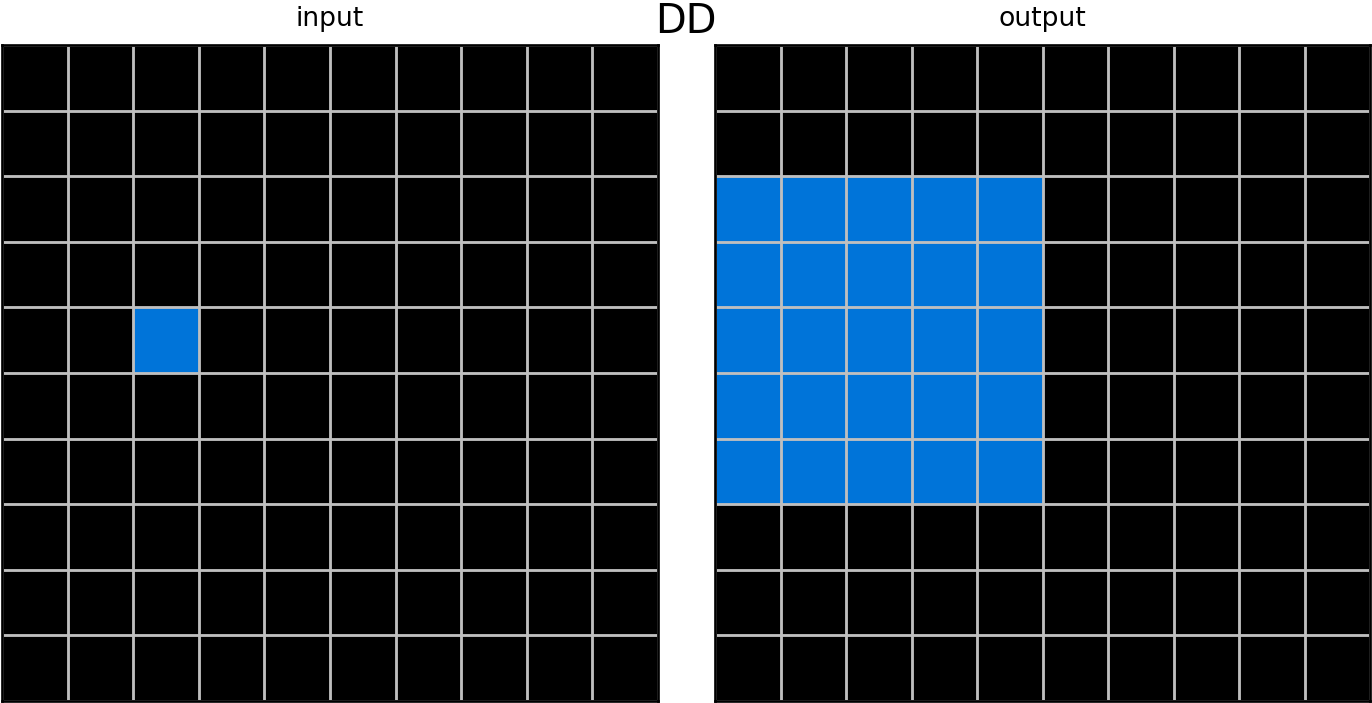}

  \vspace{0.35em}

  \includegraphics[width=0.32\textwidth]{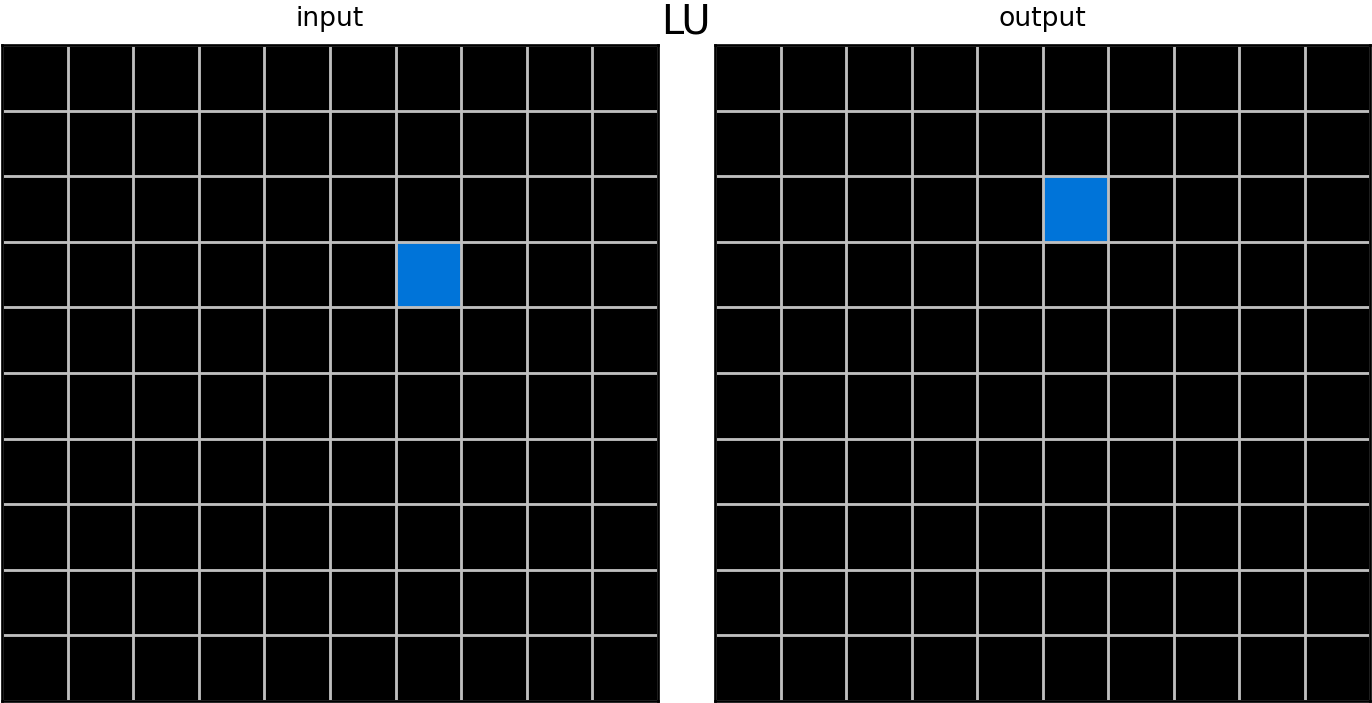}\hfill
  \includegraphics[width=0.32\textwidth]{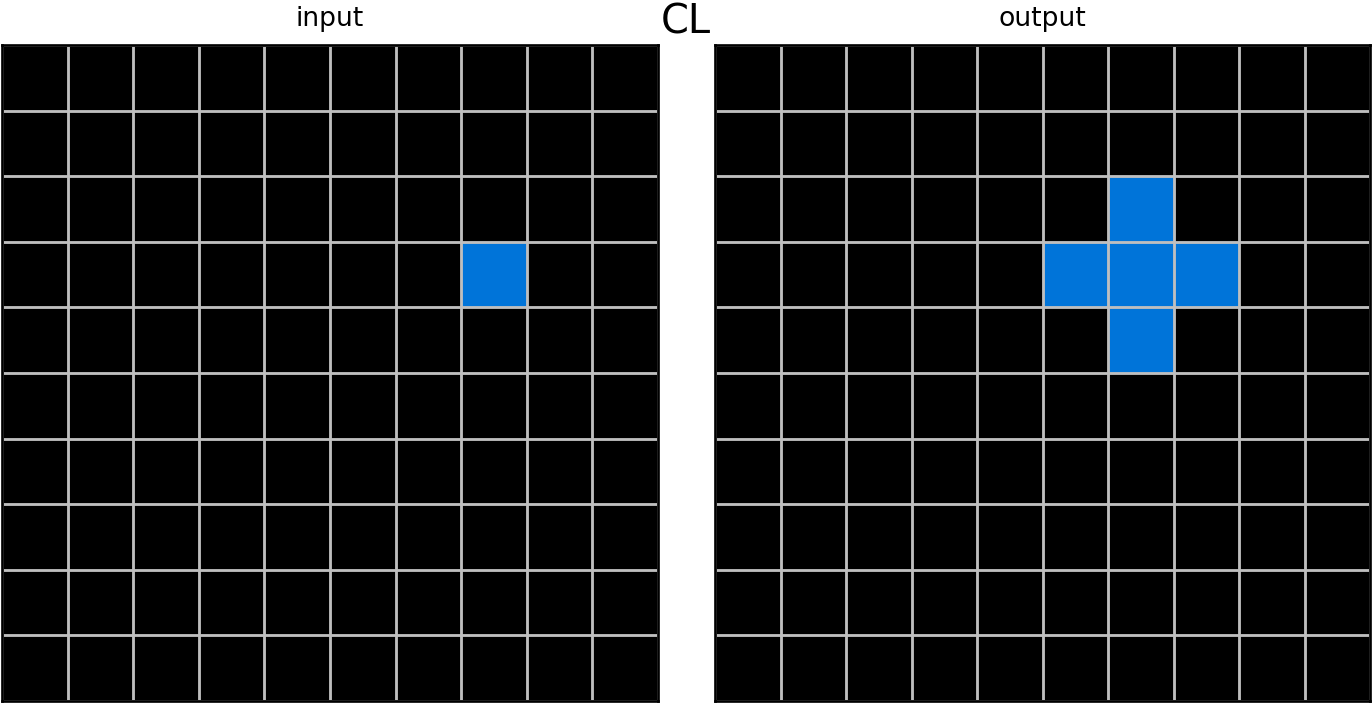}\hfill
  \includegraphics[width=0.32\textwidth]{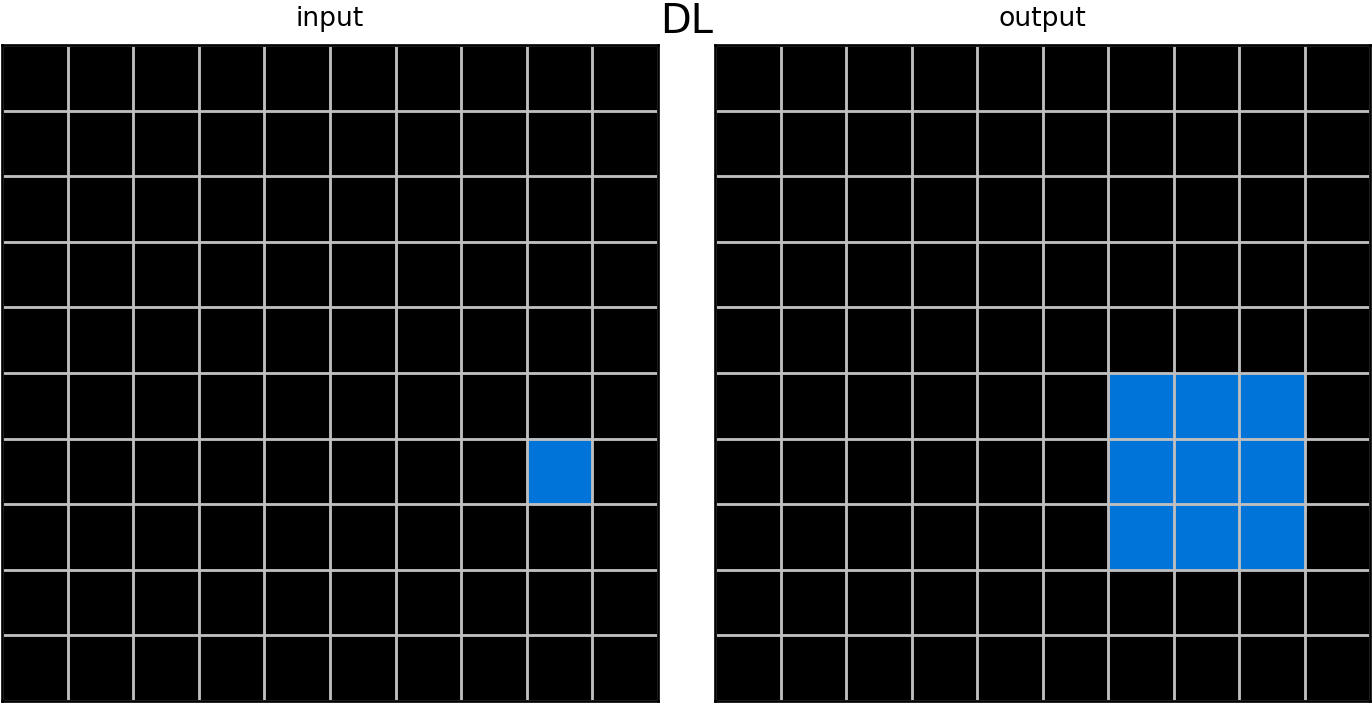}

  \vspace{0.35em}

  \includegraphics[width=0.32\textwidth]{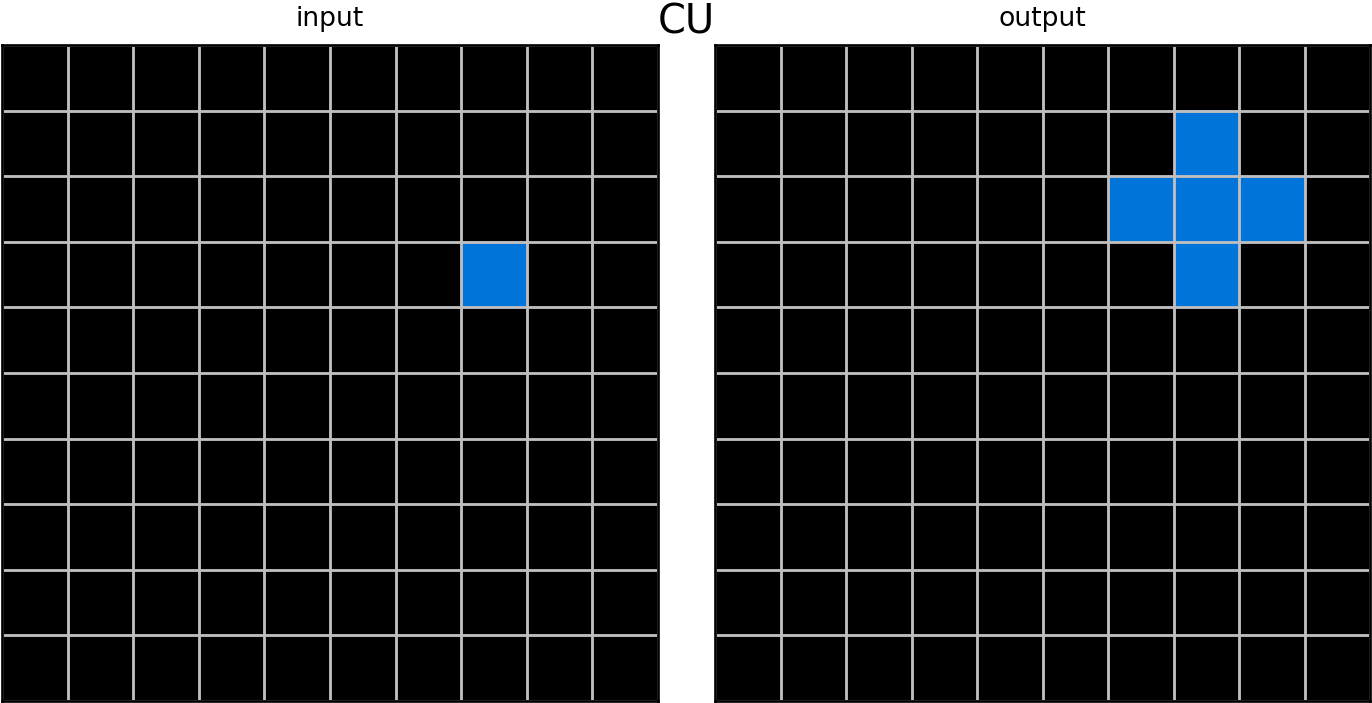}\hfill
  \includegraphics[width=0.32\textwidth]{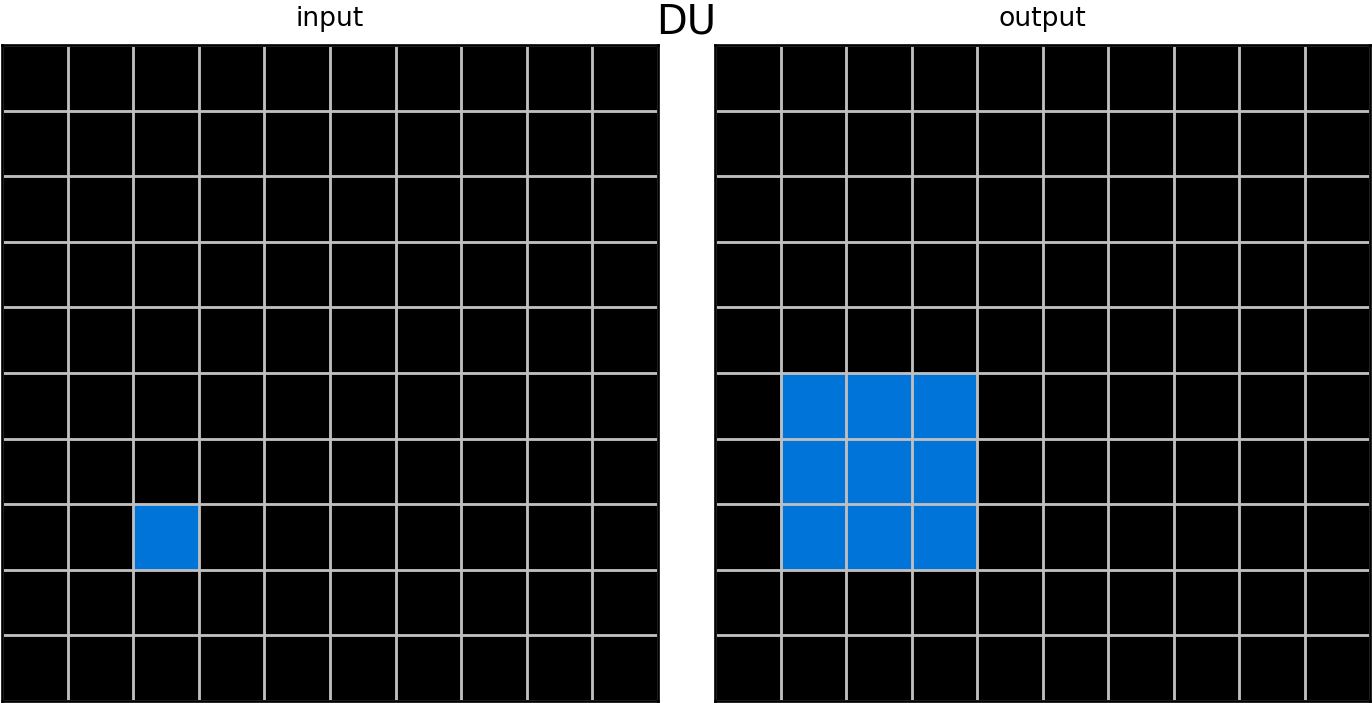}\hfill
  \includegraphics[width=0.32\textwidth]{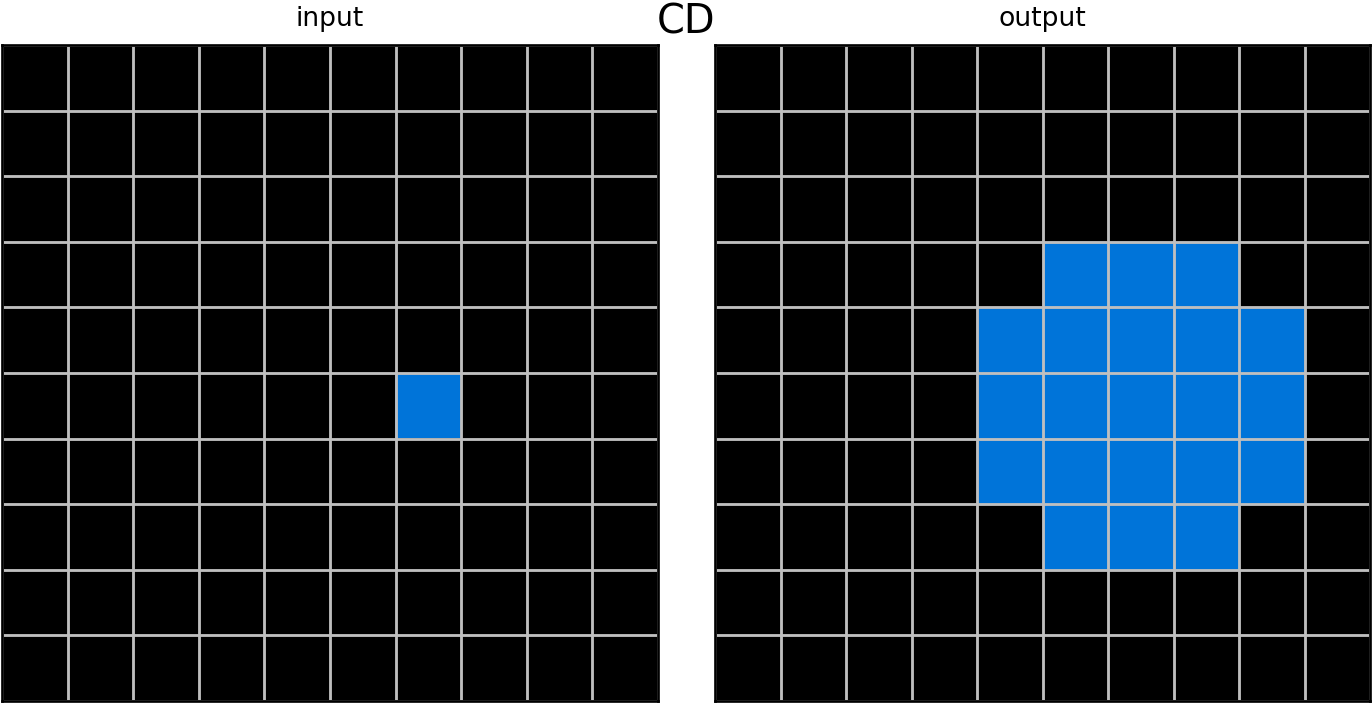}

  \caption{\textbf{Our LUCD tasks}. Examples of input-output pairs per LUCD task of our dataset. Tasks are encoded as `left' (L), `up' (U), 'cross' (C), 'dilate' (D), 'neutral' (N) transforms, composed with each other.}
  \label{fig:ludc-3x5}
\end{figure*}

One input-output pair for each of our 15 tasks of our custom LUCD dataset are represented in~\Cref{fig:ludc-3x5}.

\section{Our Custom Moves dataset}
\label{sec:moves}

\begin{figure*}[t]
  \centering
  \includegraphics[width=0.32\textwidth]{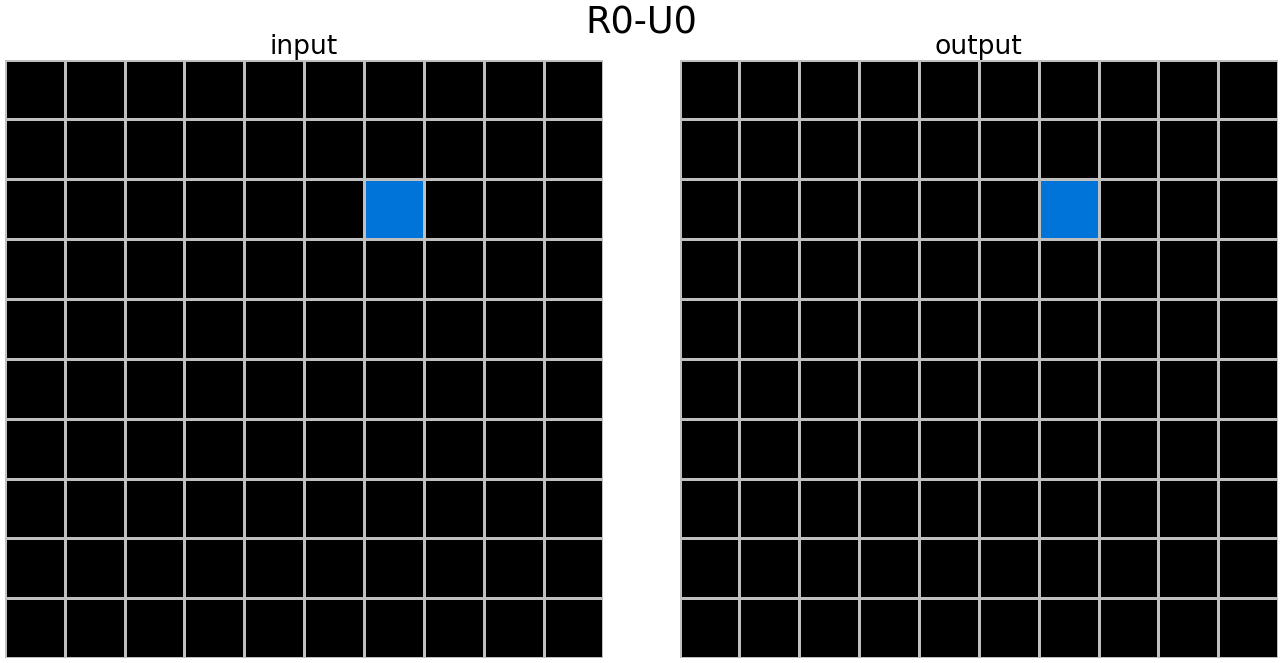}\hfill
  \includegraphics[width=0.32\textwidth]{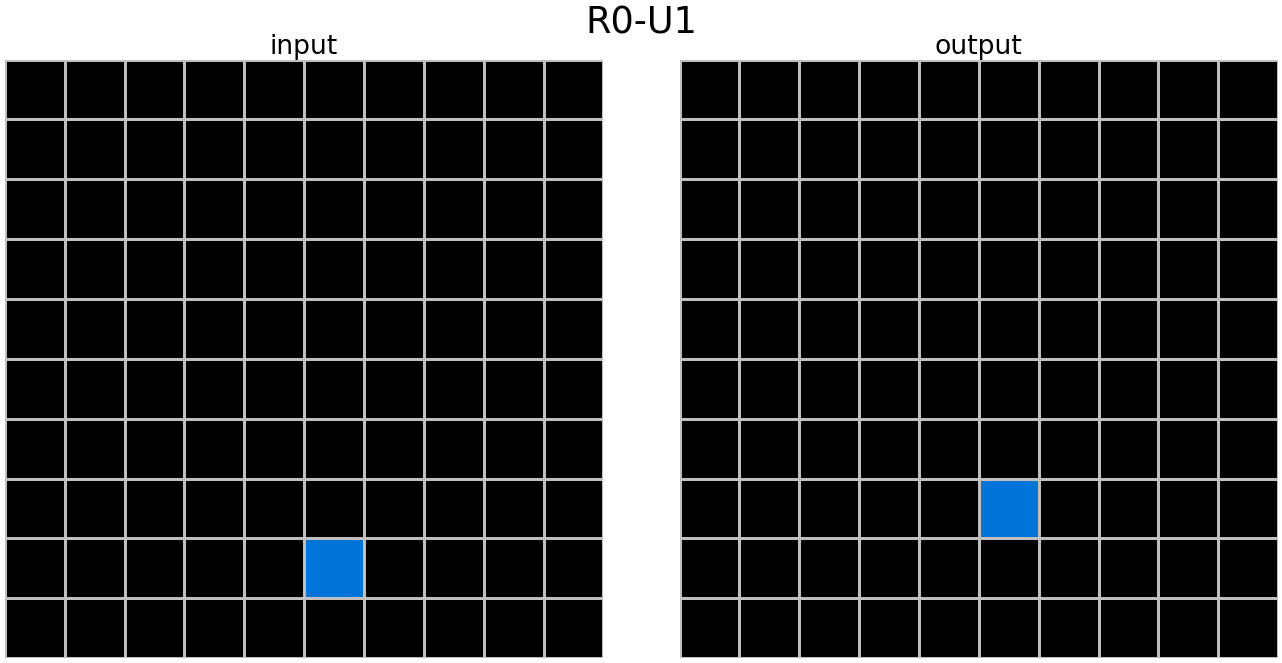}\hfill
  \includegraphics[width=0.32\textwidth]{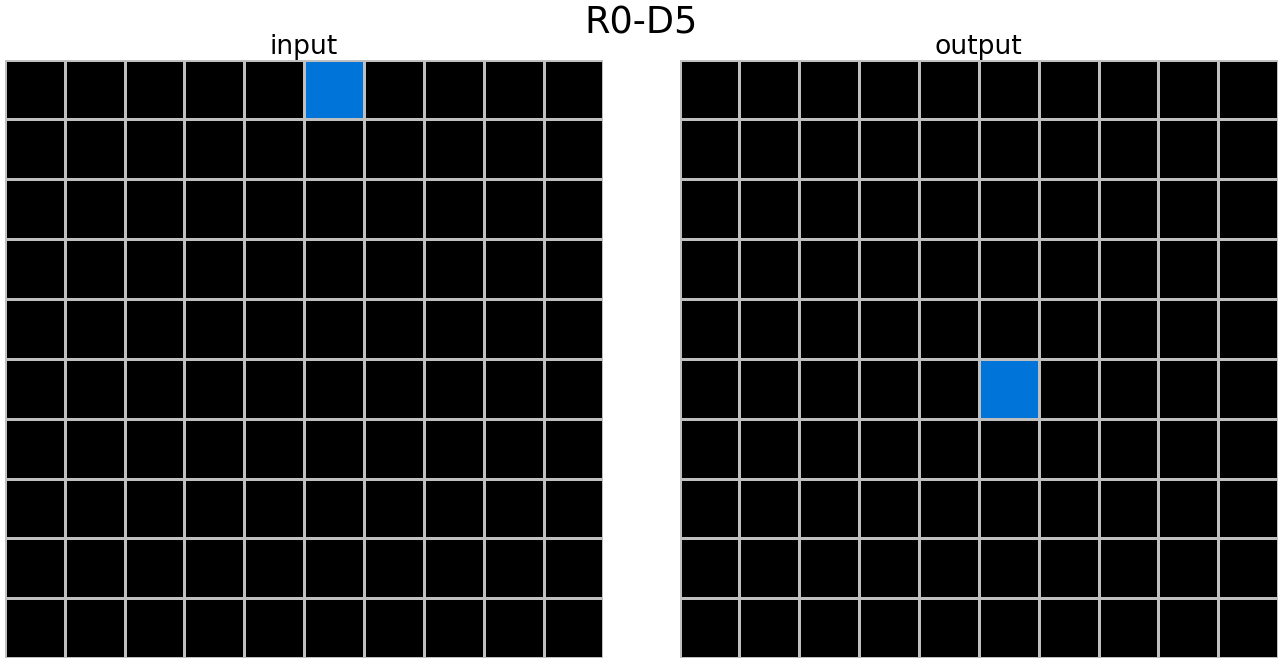}

  \vspace{0.35em}

  \includegraphics[width=0.32\textwidth]{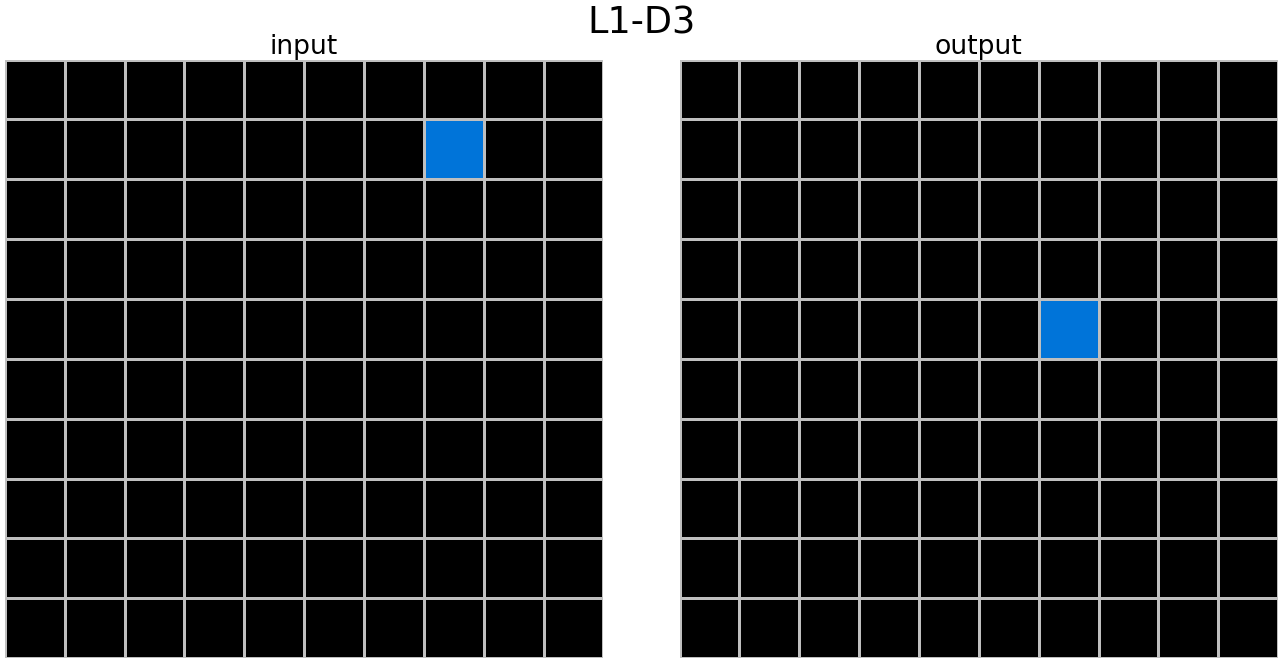}\hfill
  \includegraphics[width=0.32\textwidth]{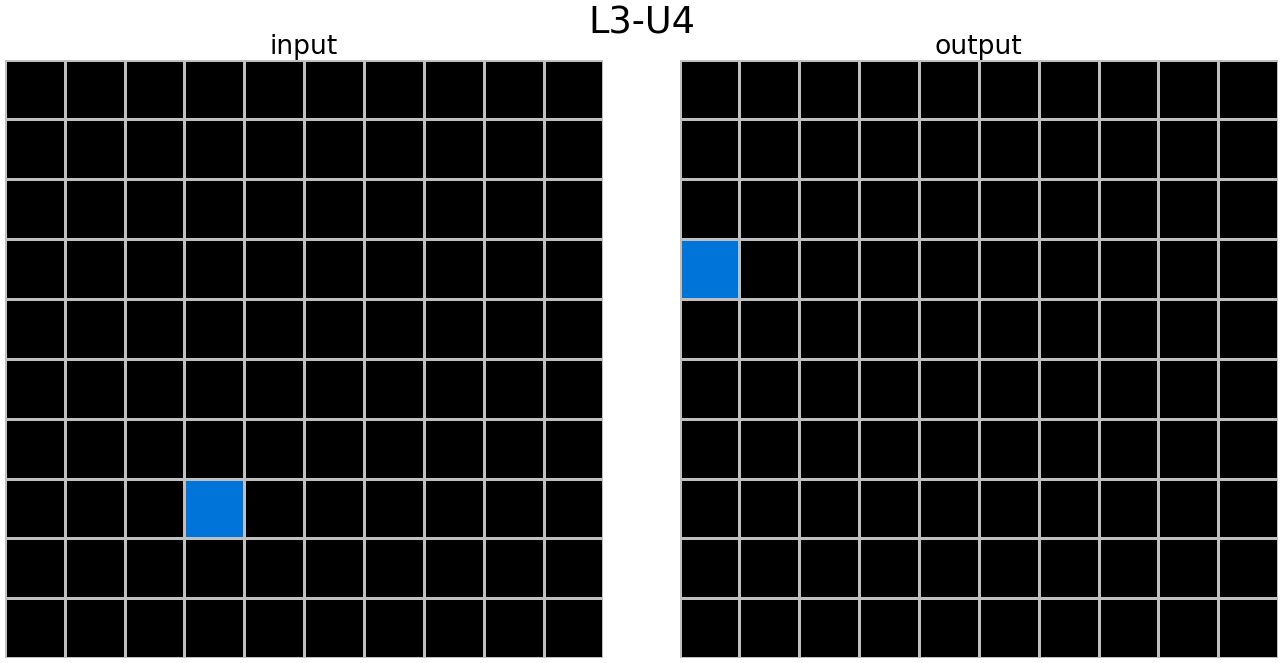}\hfill
  \includegraphics[width=0.32\textwidth]{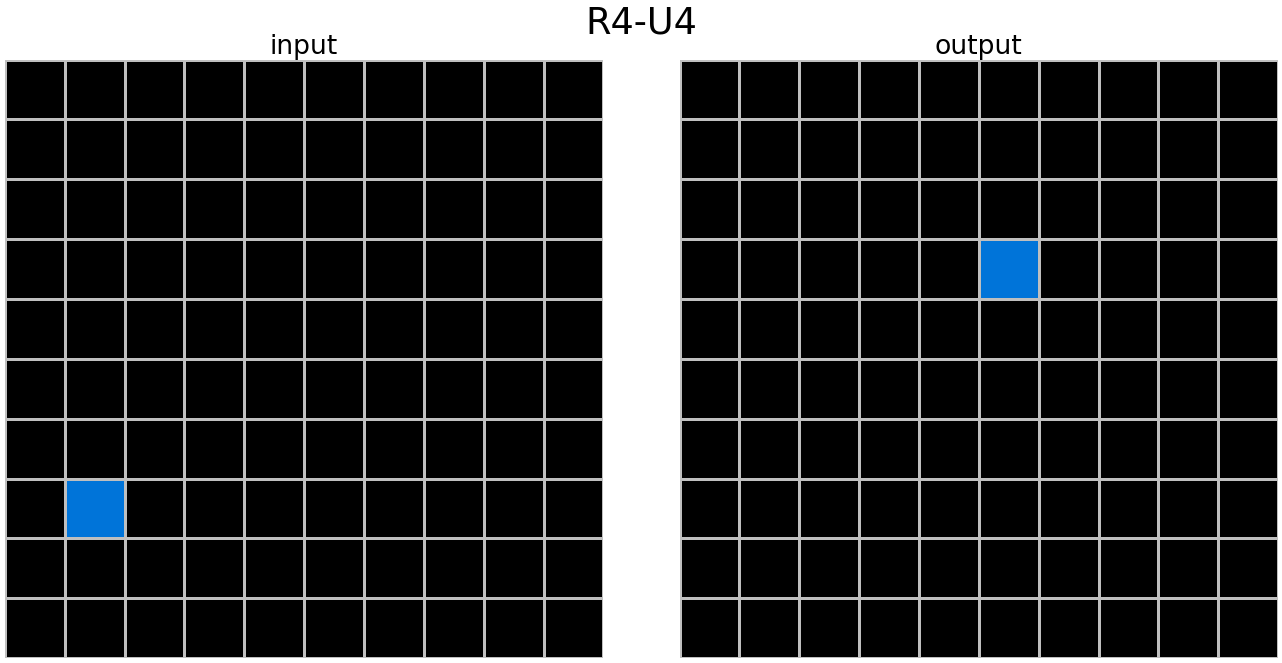}

  \caption{\textbf{Our Moves tasks.} Examples of input-output pairs per Moves task of our dataset. Tasks are encoded as left (L) or right (R) moves and up (U) and down (D) moves.}
  \label{fig:moves-2x3}
\end{figure*}

Some input-output pairs of our Moves dataset are represented in~\Cref{fig:moves-2x3}.

\section{Qualitative similar tasks retrieval}
\label{sec:retrieval-supp}

\begin{figure*}[t]
  \centering

  \includegraphics[width=0.49\textwidth]{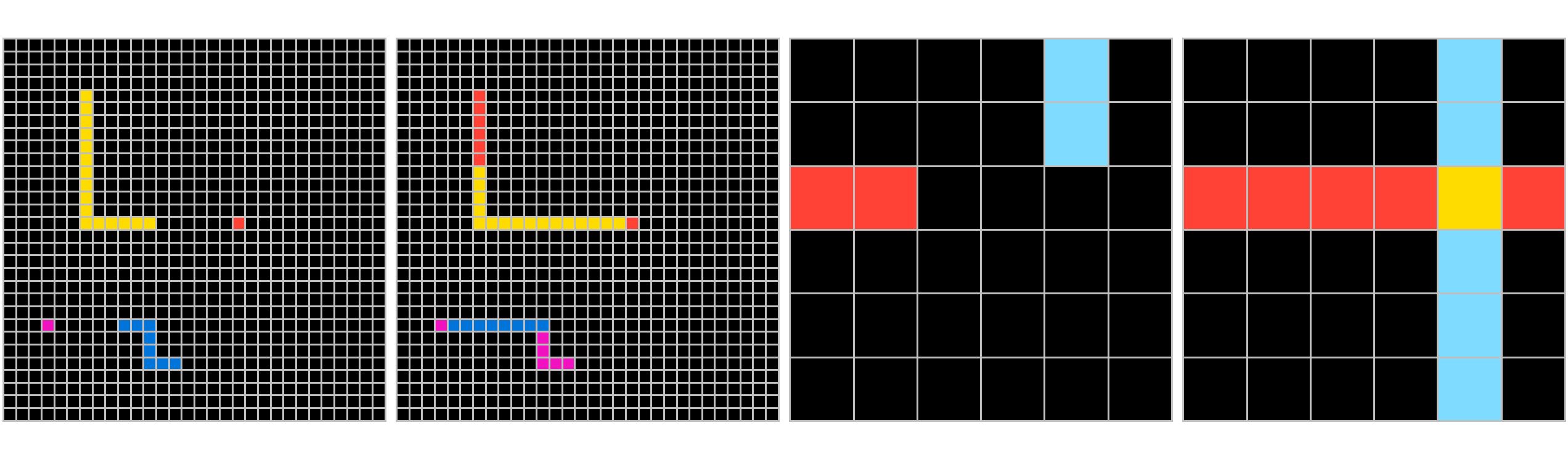}\hfill
  \includegraphics[width=0.49\textwidth]{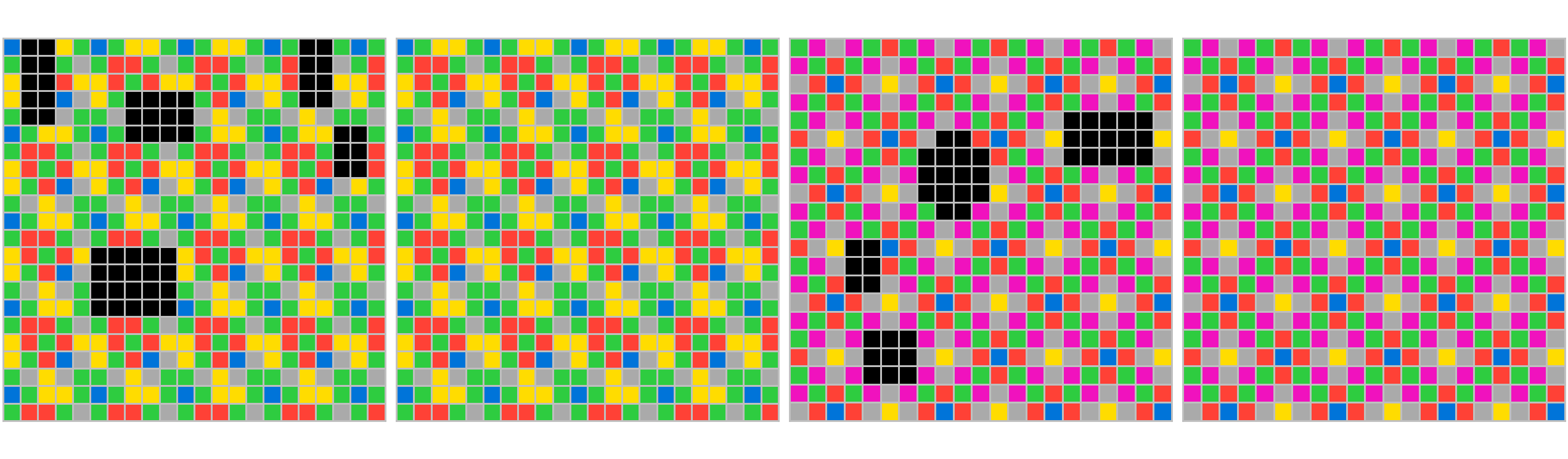}

  \vspace{0.3em}
  \includegraphics[width=0.49\textwidth]{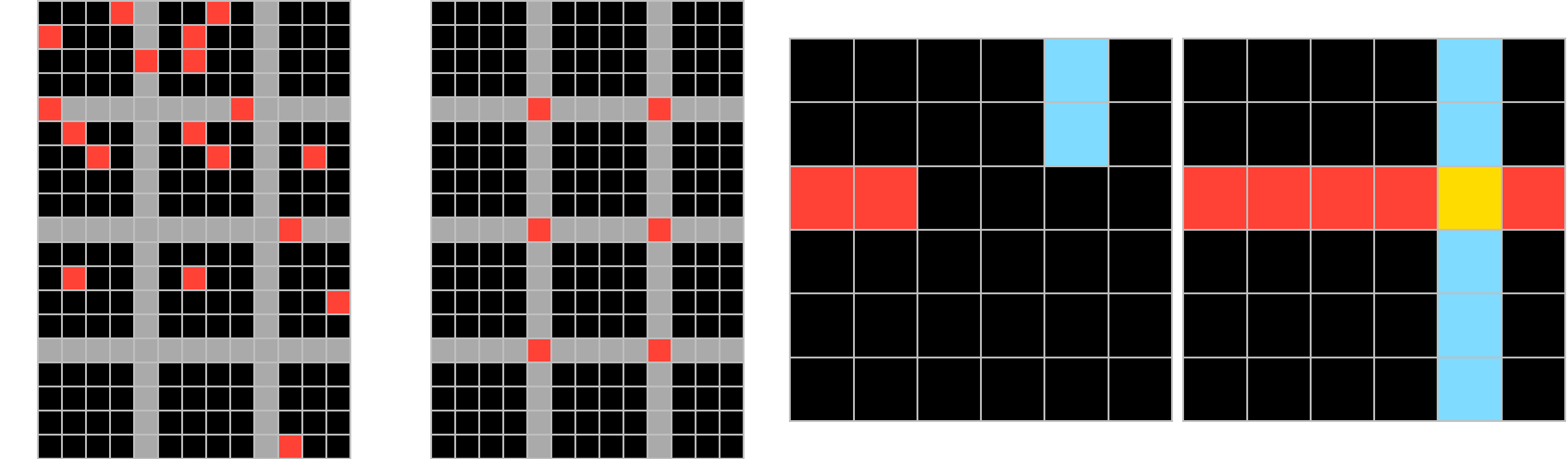}\hfill
  \includegraphics[width=0.49\textwidth]{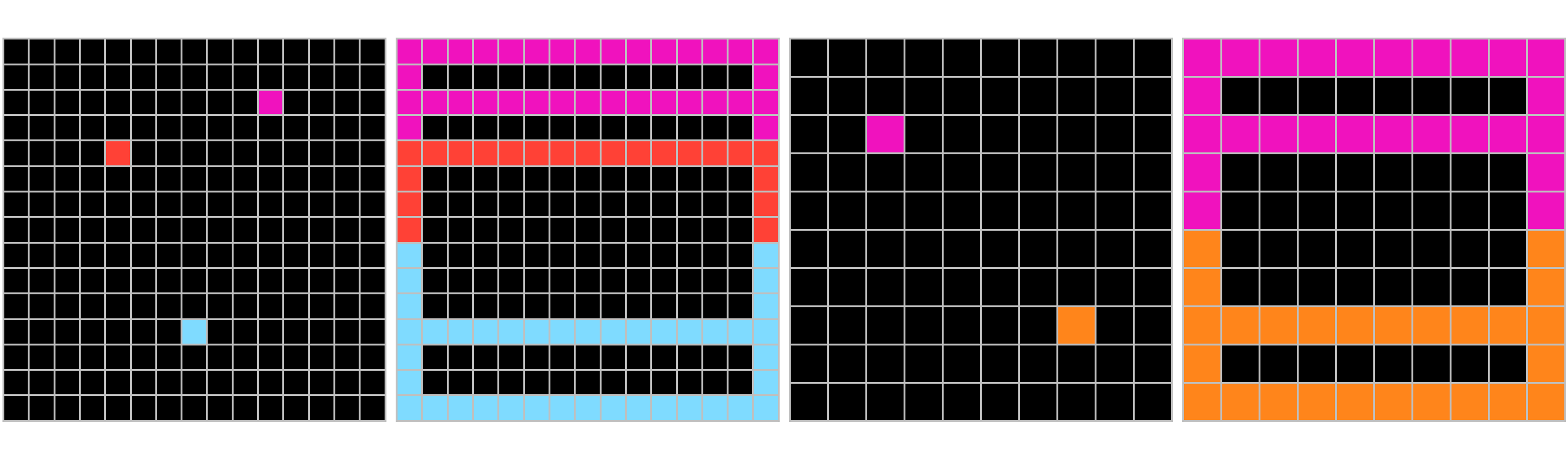}

  \vspace{0.3em}
  \includegraphics[width=0.49\textwidth]{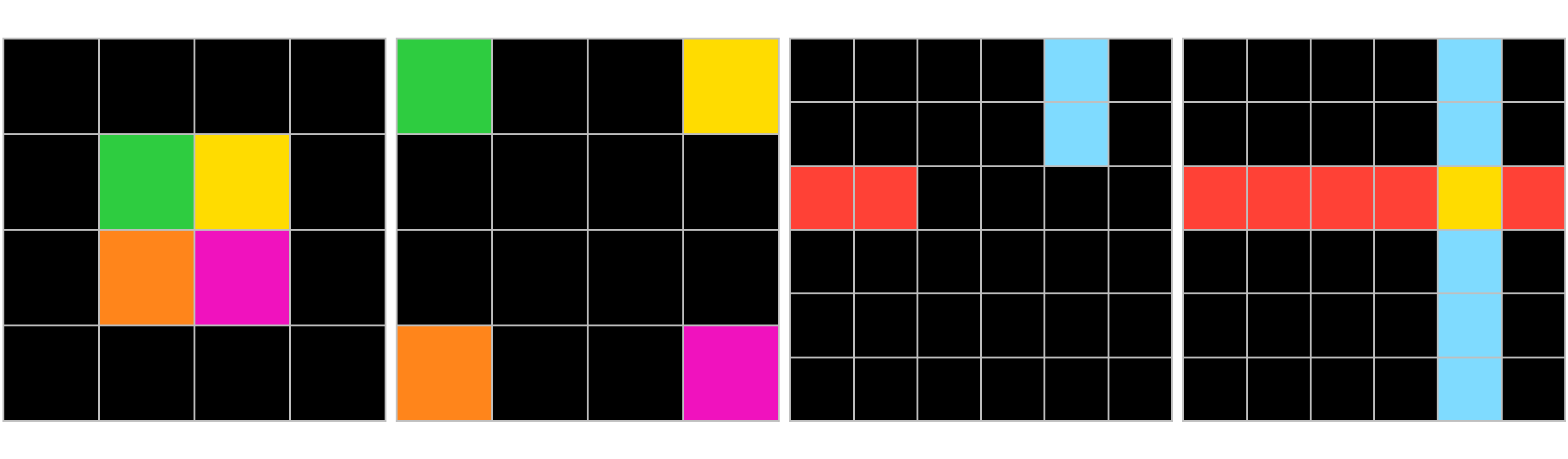}\hfill
  \includegraphics[width=0.49\textwidth]{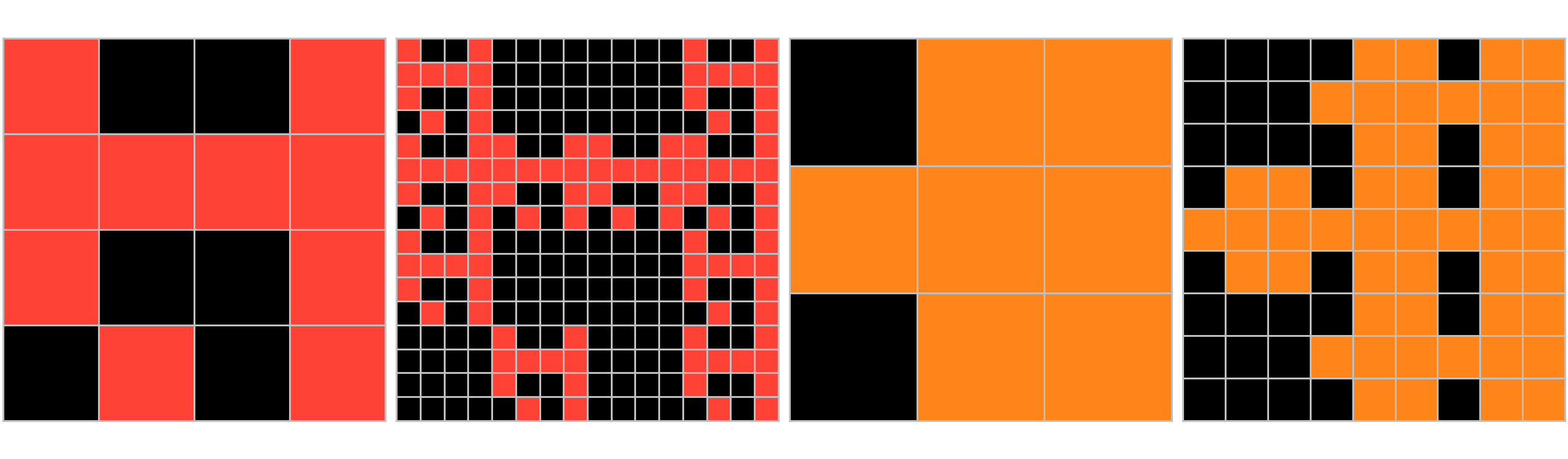}

  \vspace{0.3em}
  \includegraphics[width=0.49\textwidth]{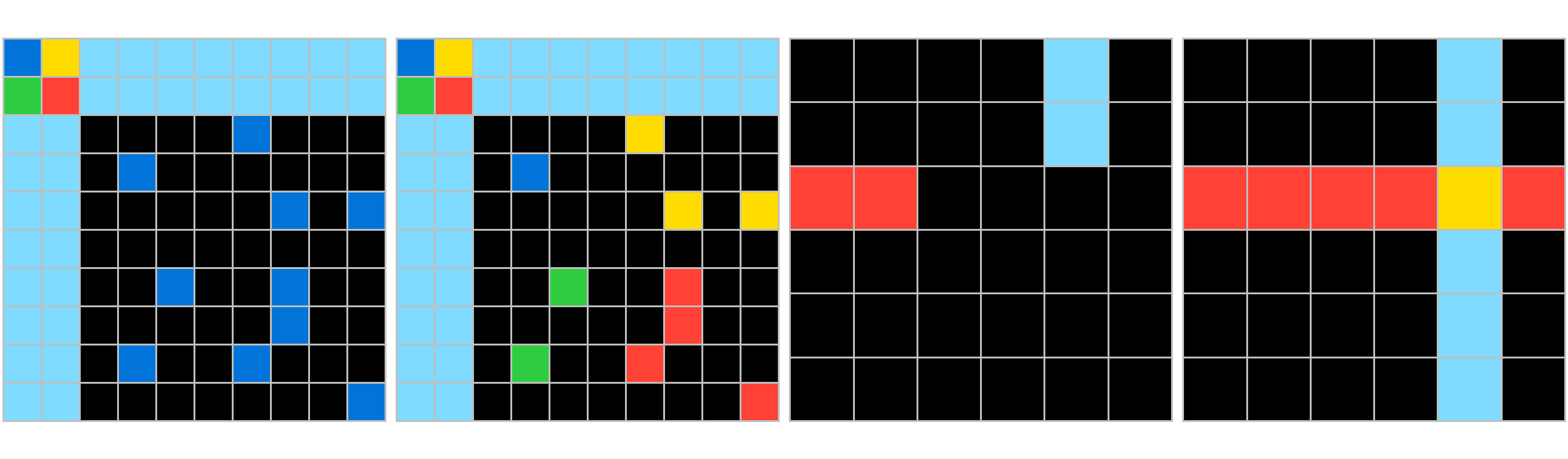}\hfill
  \includegraphics[width=0.49\textwidth]{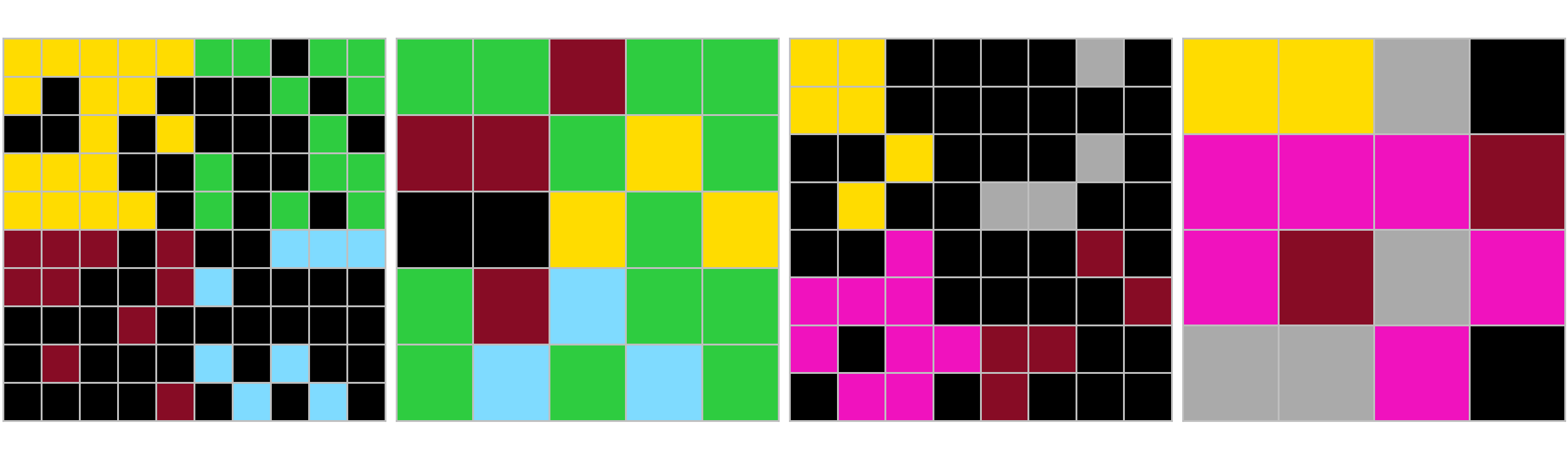}

  \vspace{0.3em}
  \includegraphics[width=0.49\textwidth]{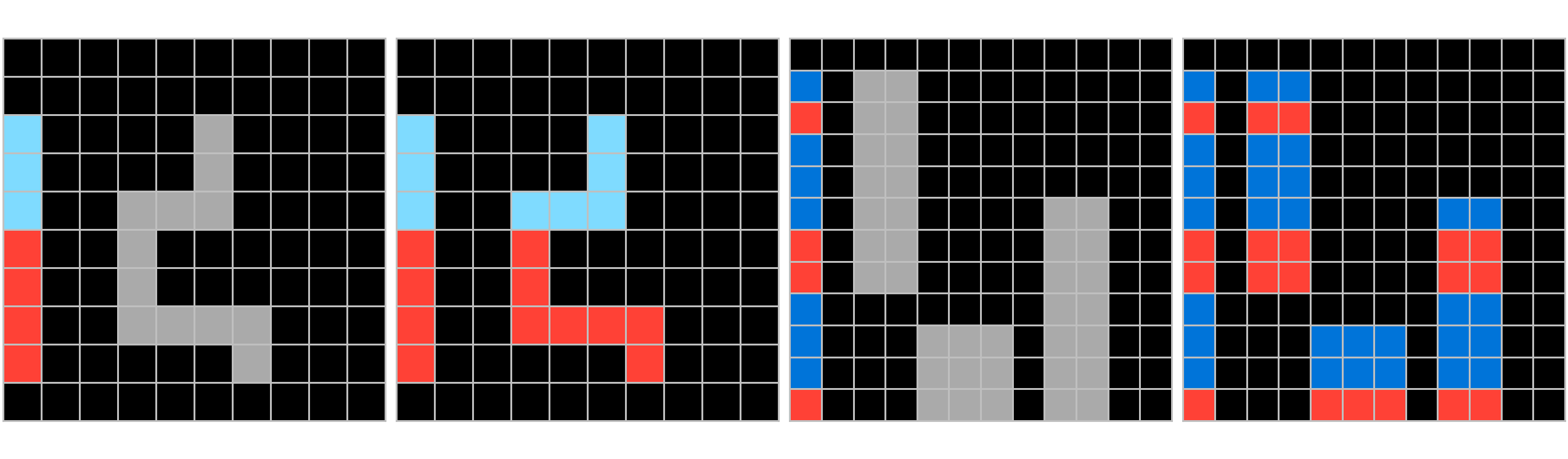}\hfill
  \includegraphics[width=0.49\textwidth]{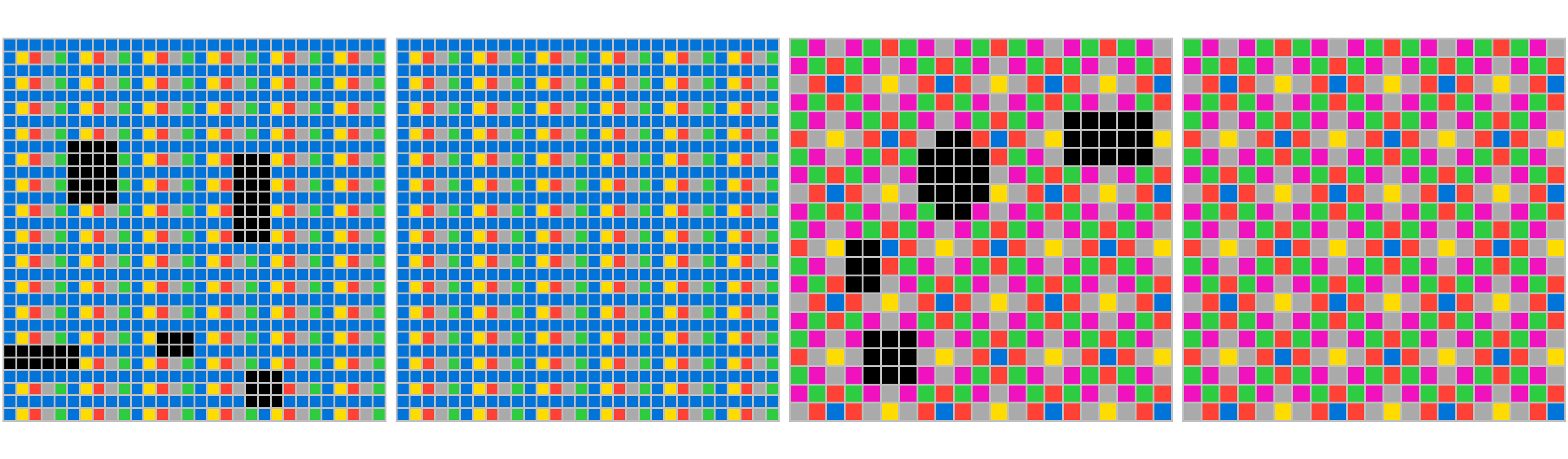}

  \vspace{0.3em}
  \includegraphics[width=0.49\textwidth]{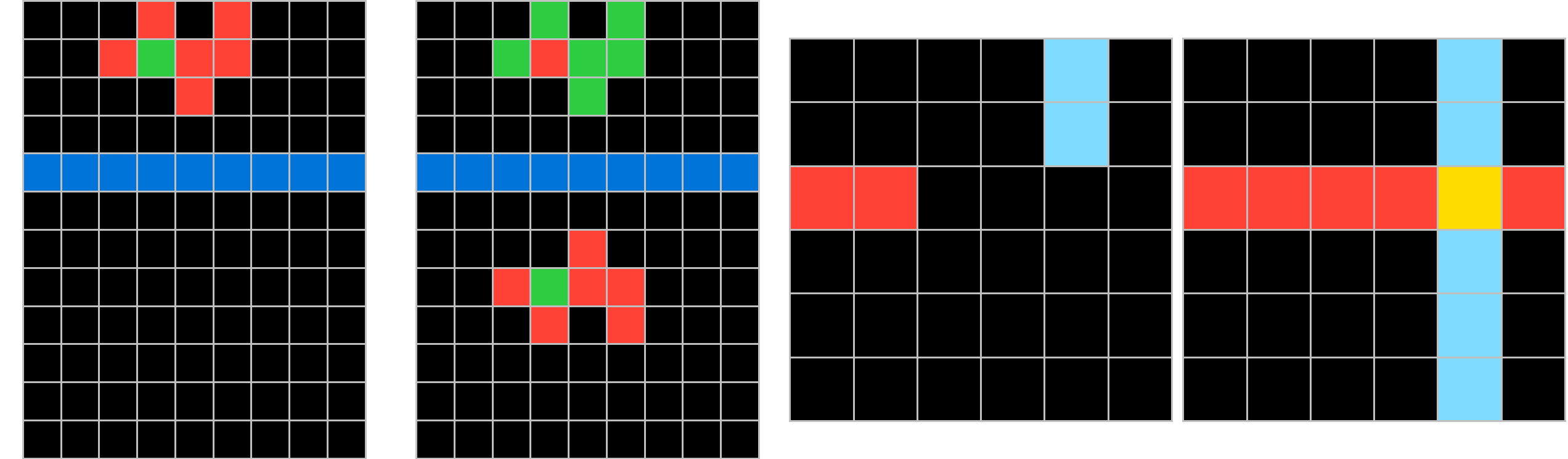}\hfill
  \includegraphics[width=0.49\textwidth]{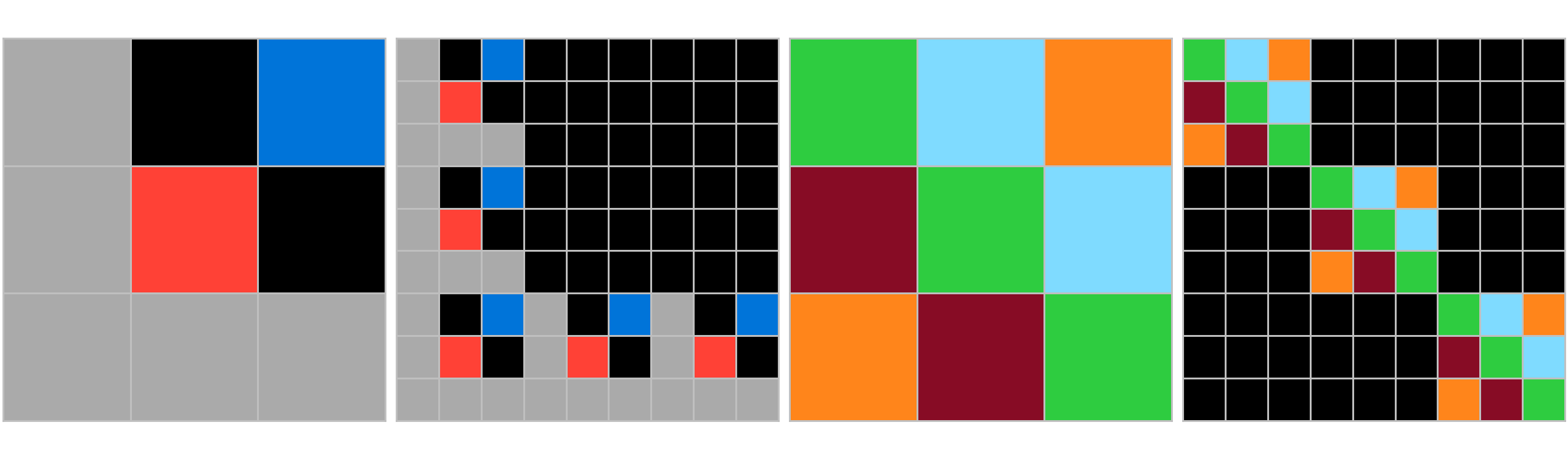}

  \vspace{0.3em}
  \includegraphics[width=0.49\textwidth]{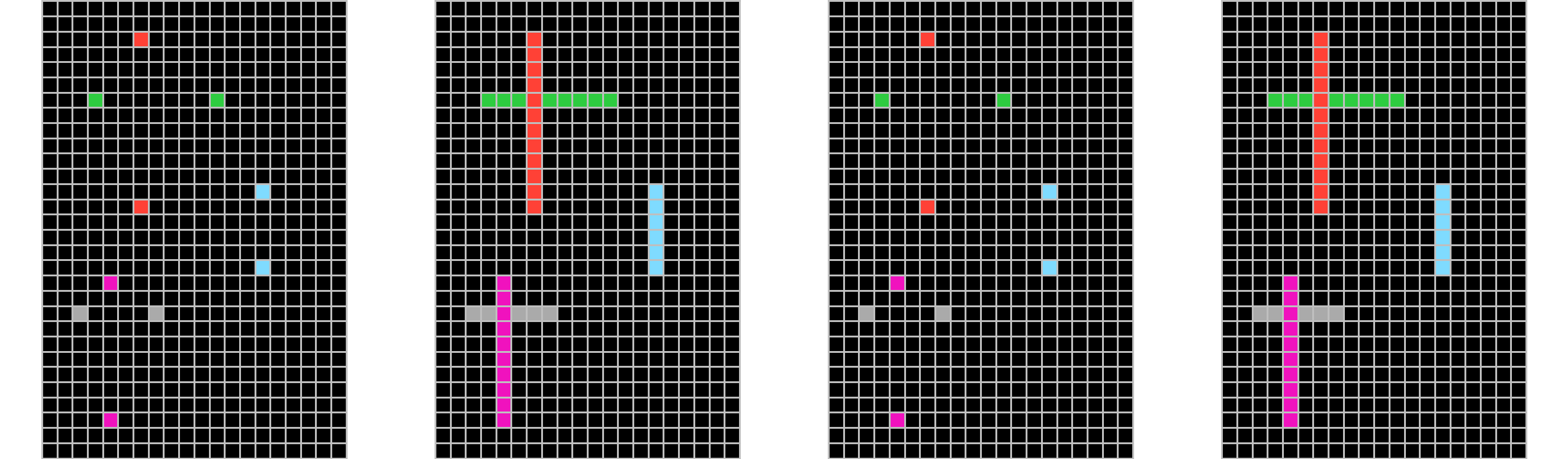}\hfill
  \includegraphics[width=0.49\textwidth]{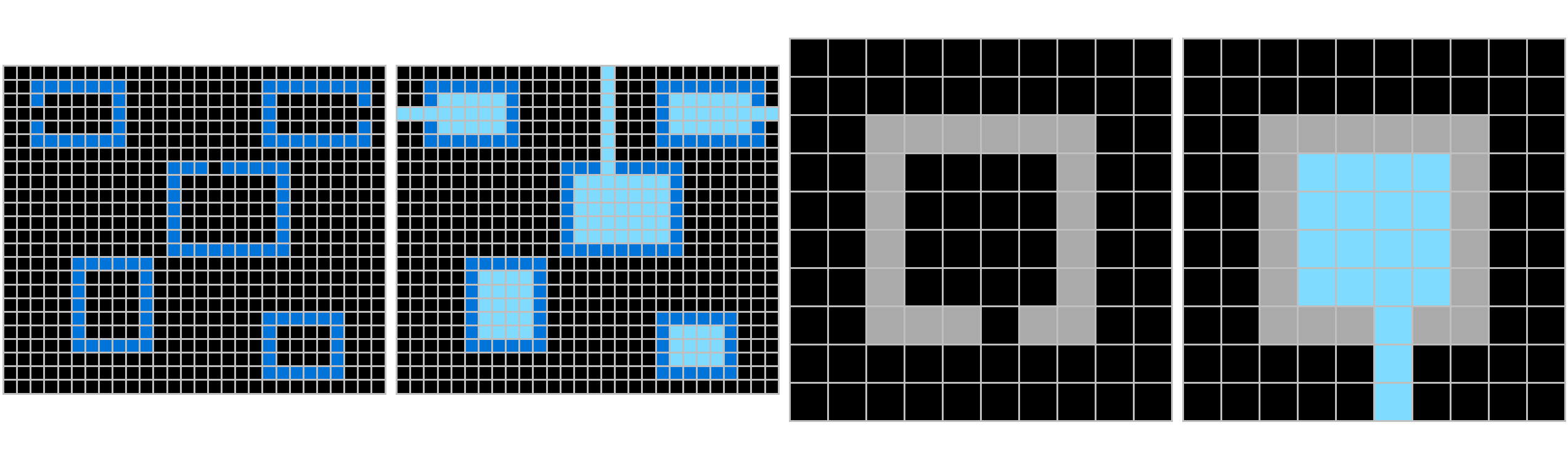}

  \vspace{0.3em}
  \includegraphics[width=0.49\textwidth]{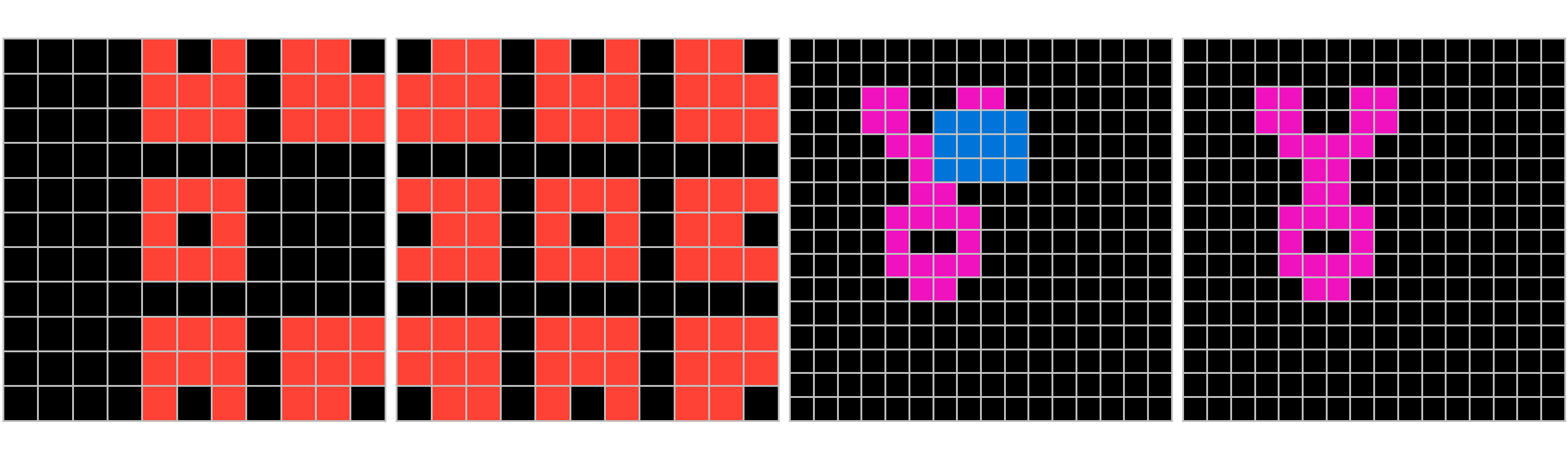}\hfill
  \includegraphics[width=0.49\textwidth]{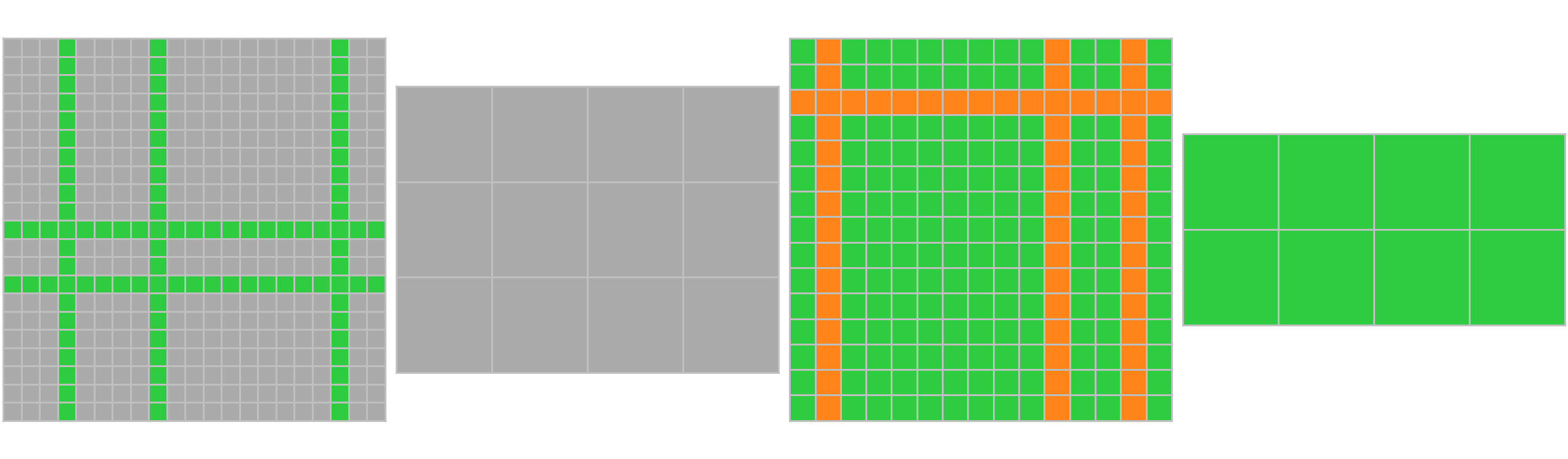}

  \vspace{0.3em}
  \includegraphics[width=0.49\textwidth]{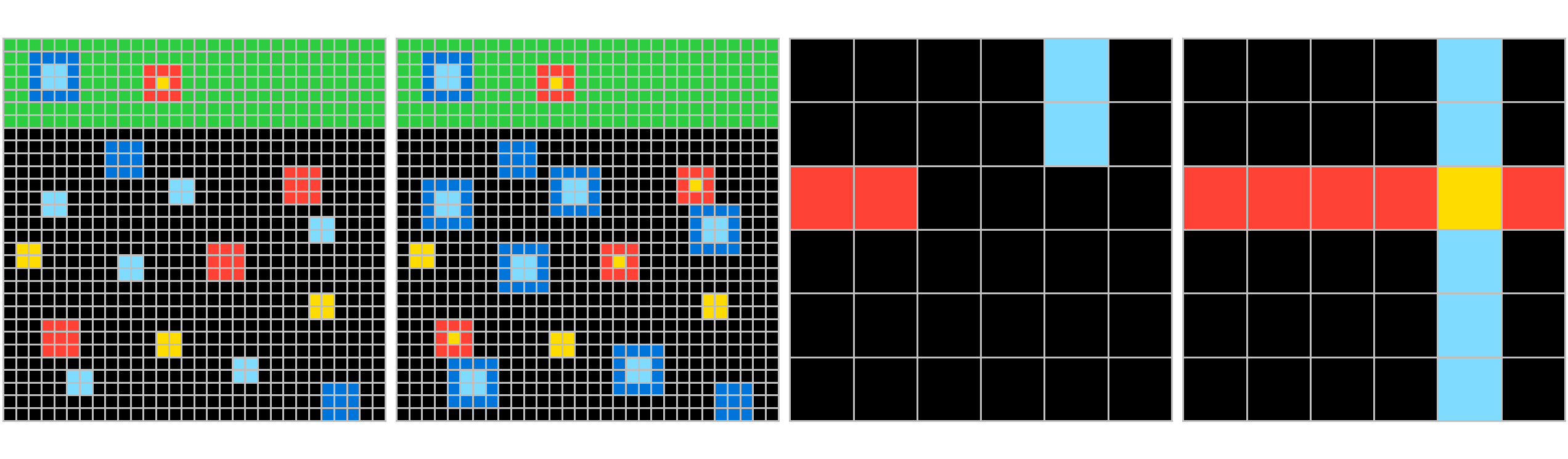}\hfill
  \includegraphics[width=0.49\textwidth]{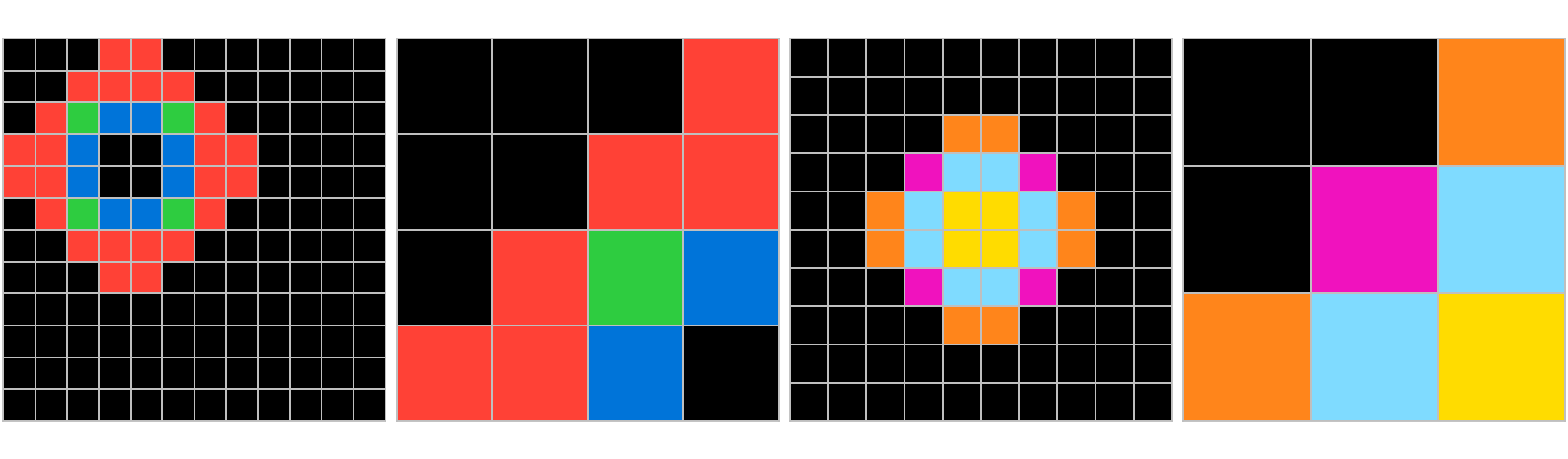}

  \vspace{0.3em}
  \includegraphics[width=0.49\textwidth]{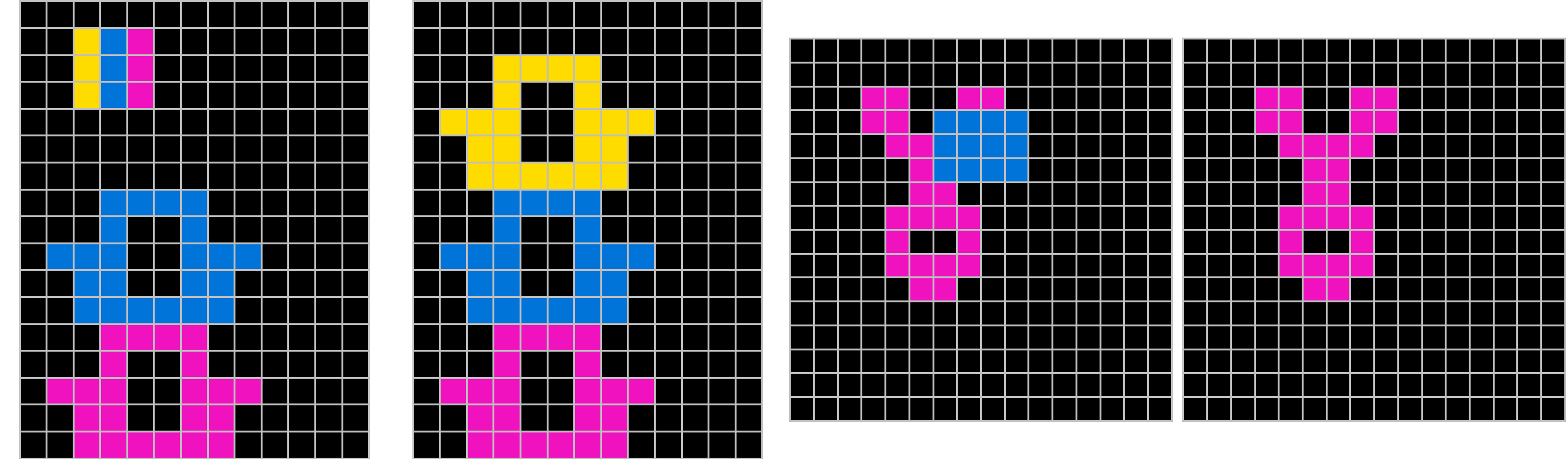}\hfill
  \includegraphics[width=0.49\textwidth]{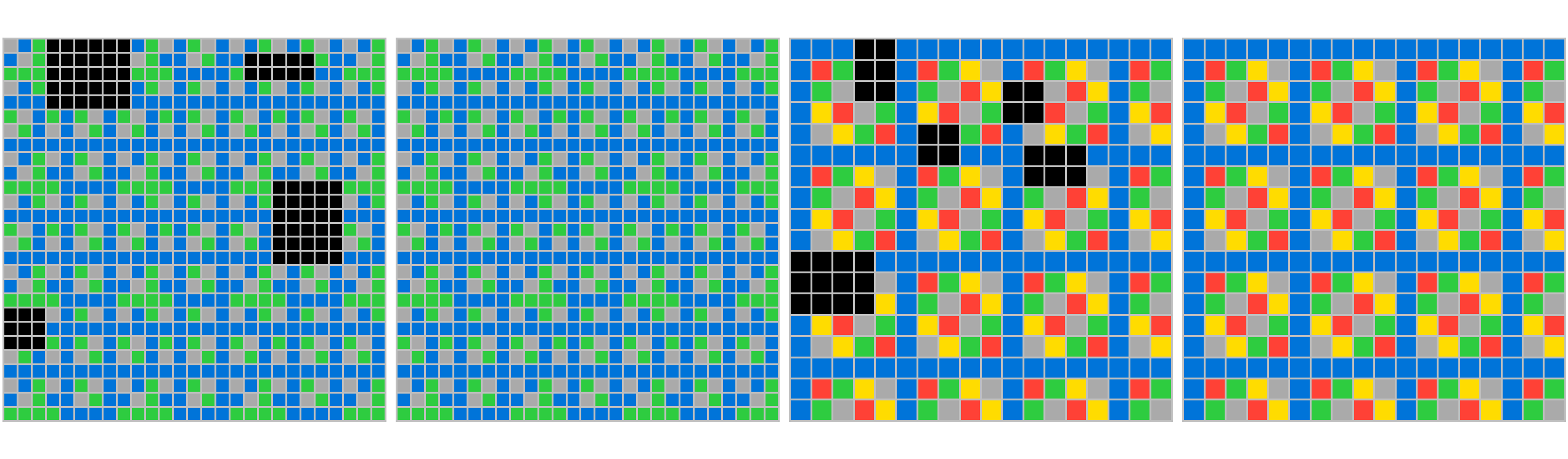}

  \caption{\textbf{Similarity retrieval.} (Left) \defaultTTT retrieves test-train pairs of tasks that do not seem semantically similar. (Right) \ourTTT retrieves test-train pairs of tasks that share some common rules.}
  \label{fig:task-similarity-supp}
\end{figure*}

Additional results of the retrieval of train tasks 'similar' to test tasks according to \defaultTTT and \ourTTT are shown in~\Cref{fig:task-similarity-supp}.

\section{PCA results on Moves}
\label{sec:pca-moves}

\begin{figure*}[t]
  \centering
  \includegraphics[width=0.7\textwidth]{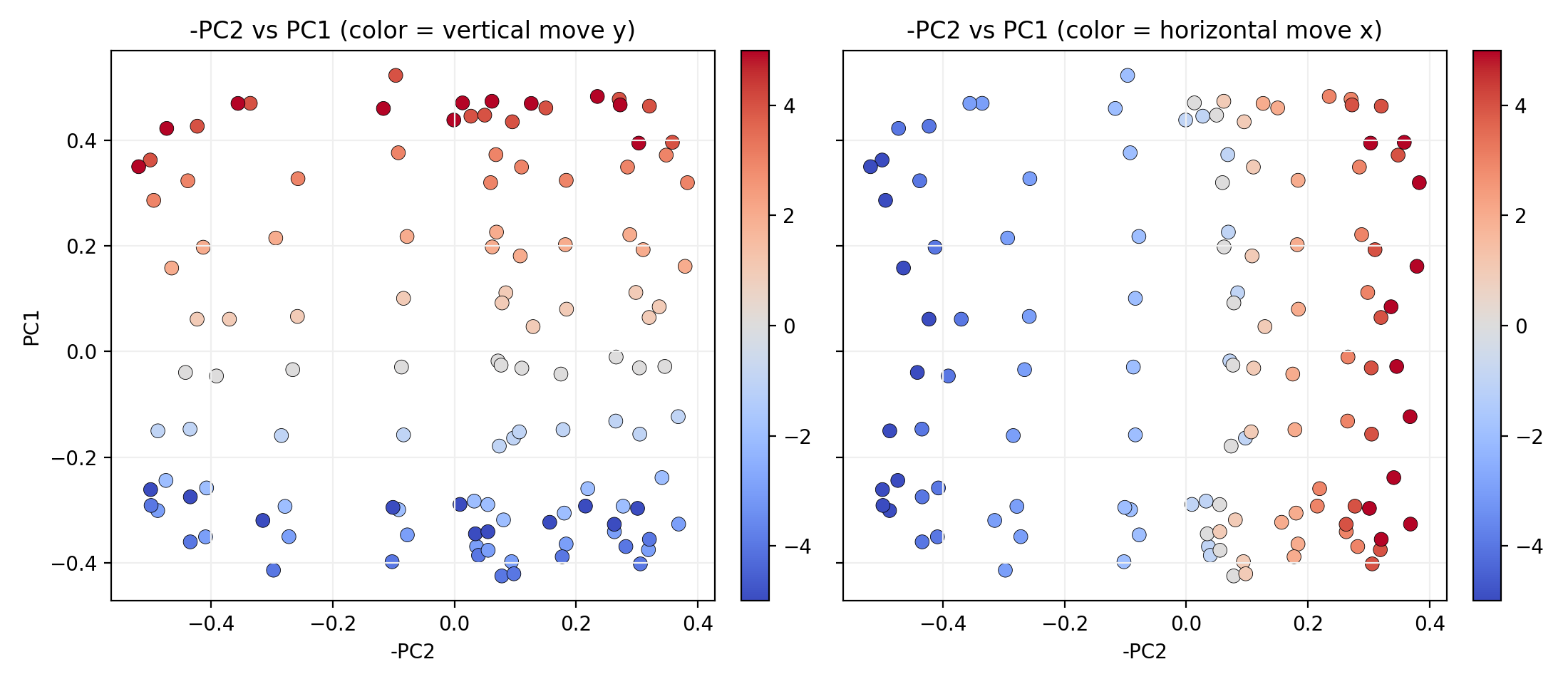}
  \includegraphics[width=0.7\textwidth]{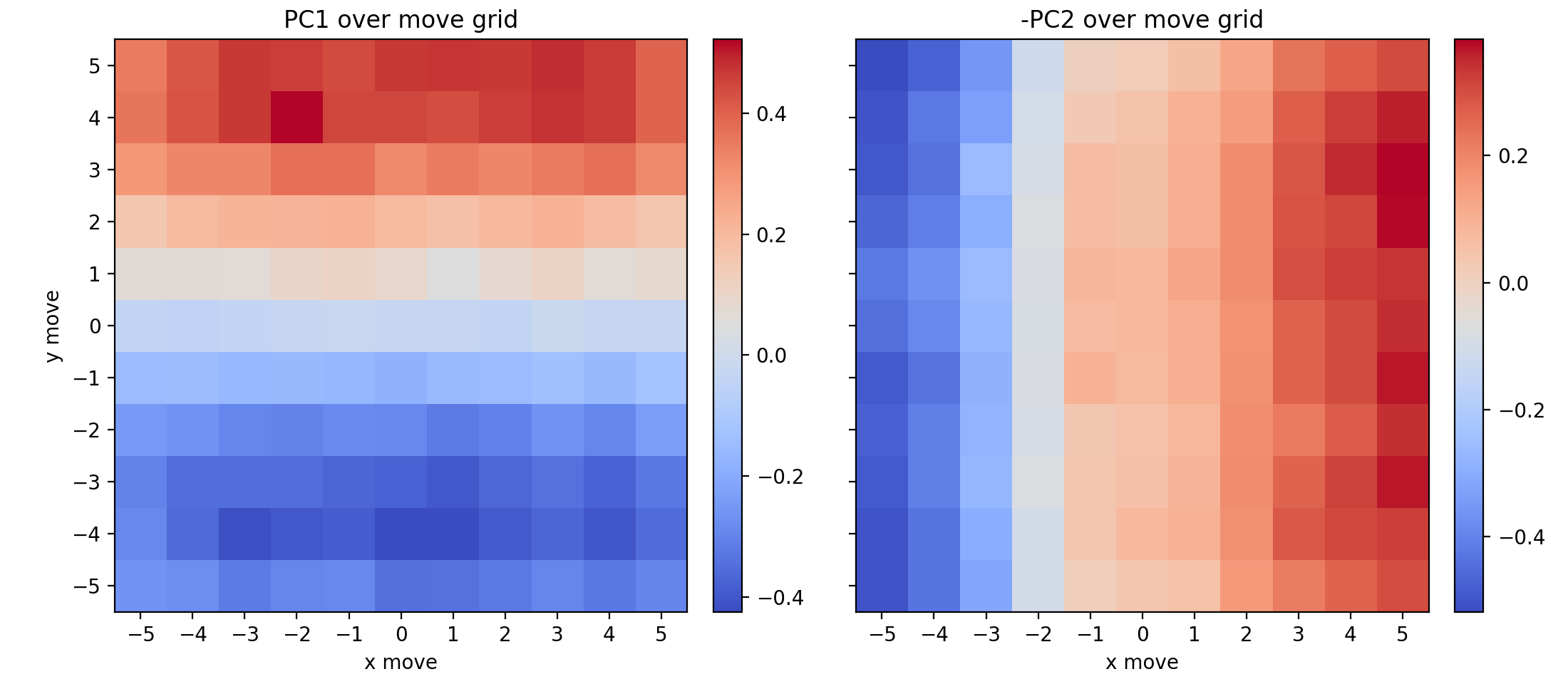}
  \caption{\textbf{\ourTTT PCA on our Moves tasks.} (Top) PCA visualization of the \ourTTT embeddings of our Moves tasks, colored according to ground truth vertical (left) and horizontal (right) moves. (Bottom) Heatmap showing, for each ground-truth move (represented as a square at (x,y) coordinates), its two main PCA values. These two equivalent plots show that the first component (PCA1) captures vertical moves, the second component (PCA2) captures horizontal moves.}
  \label{fig:pca-moves}
\end{figure*}

\begin{figure*}[t]
  \centering
  \includegraphics[width=0.7\textwidth]{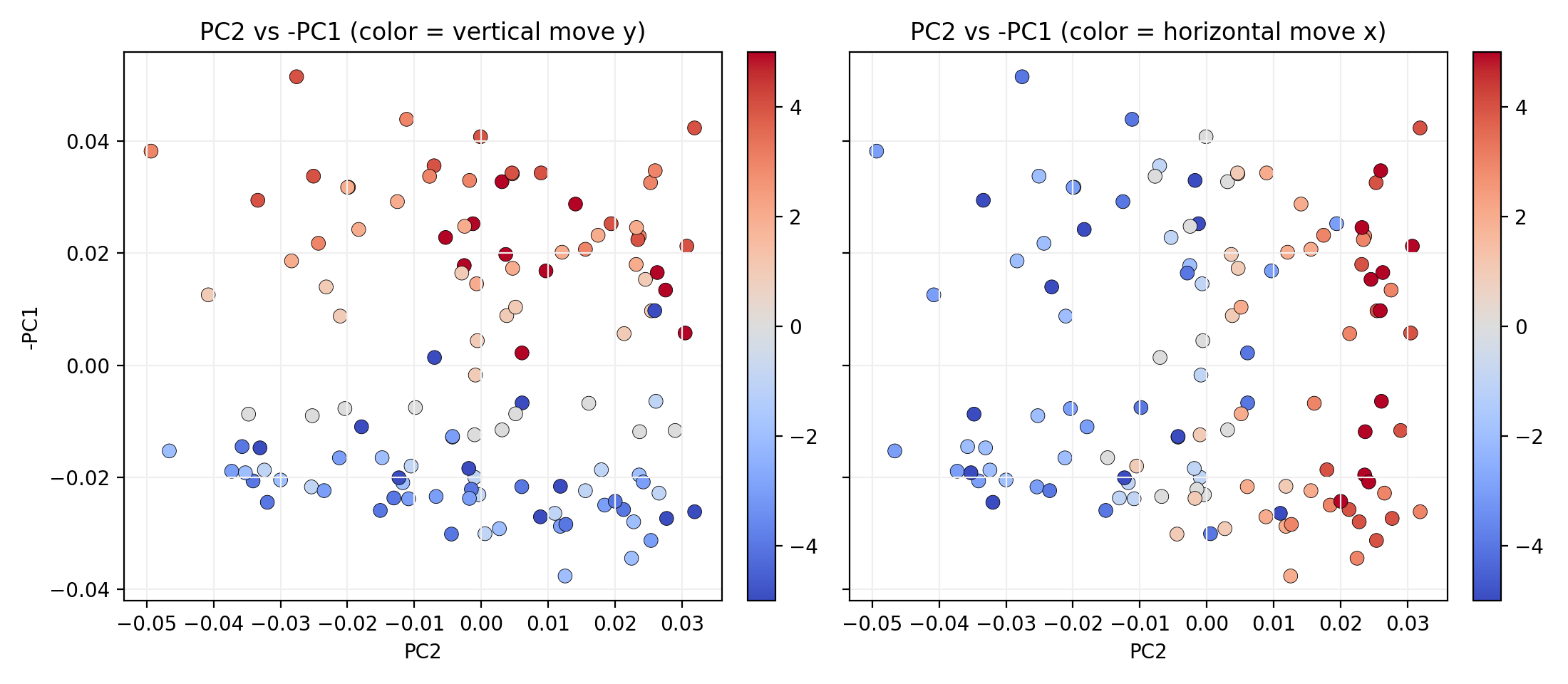}
  \includegraphics[width=0.7\textwidth]{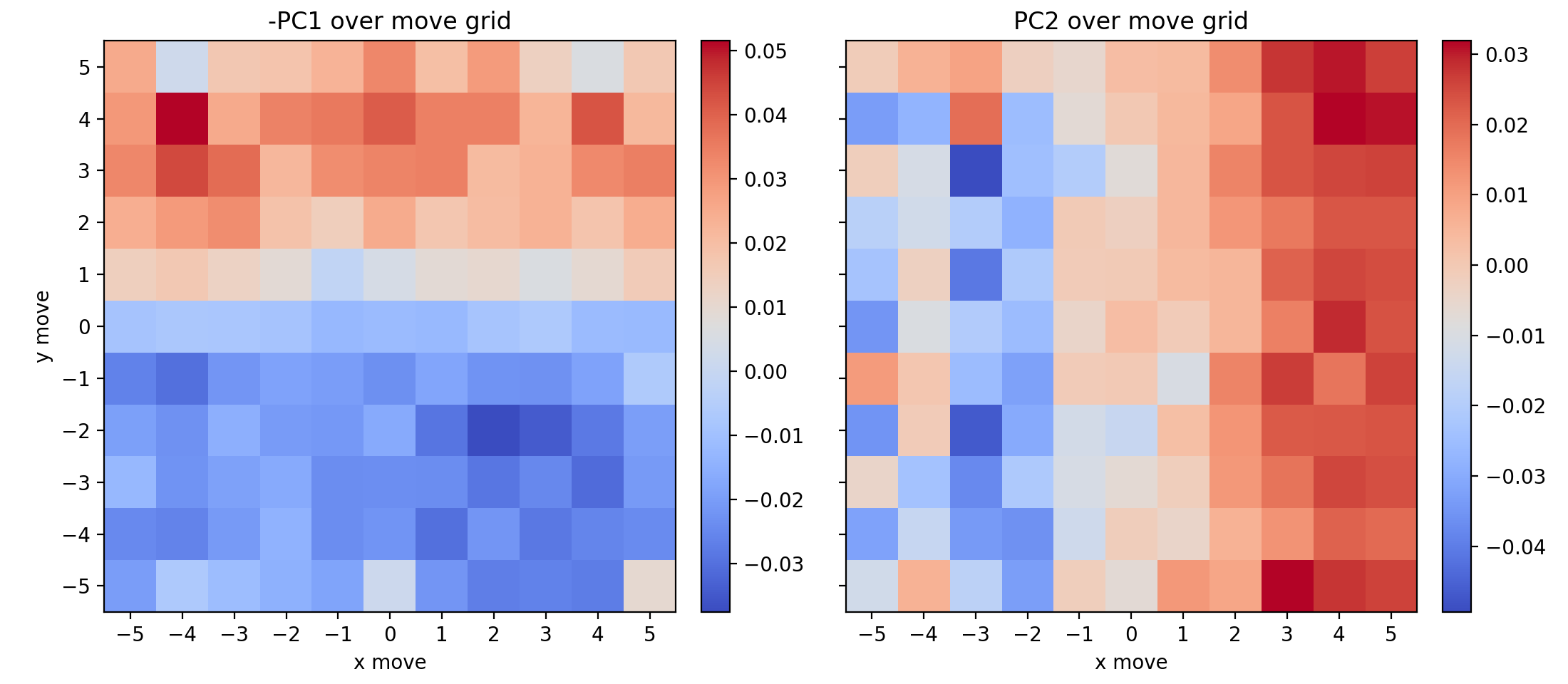}
  \caption{\textbf{\defaultTTT PCA on our Moves tasks.} (Top) PCA visualization of the \defaultTTT embeddings of our Moves tasks, colored according to ground truth vertical (left) and horizontal (right) moves. (Bottom) Heatmap showing, for each ground-truth move (represented as a square at (x,y) coordinates), its two main PCA values. \defaultTTT seems to distinguish left moves from right moves, and up moves from down moves, but the exact move ordering seems much less recaptured than with \ourTTT.}
  \label{fig:pca-moves-fullTTT}
\end{figure*}

\ourTTT embeddings capture well vertical and horizontal ground-truth moves, as shown in~\Cref{fig:pca-moves}. In comparison, \defaultTTT mostly captures the left-right and up-down distinctions, but not a fine regular ordering of the moves, as shown in~\Cref{fig:pca-moves-fullTTT}. The PCA values also tend to indicate a much more condensed projection.

\section{Experimenting with moves and color switch}
\label{sec:move_switch}

We designed a controlled \ARClike benchmark where each task applies (i) a spatial move and (ii) a color switch to a single blue cell. Spatial moves span displacements from \(-3\) to \(+3\) on each axis (\(7\times7=49\) move rules), and color switches span 8 target colors, yielding \(49\times 8=392\) task types. 
The training split contains only \emph{partial} rules: \emph{move-only} tasks (all moves with no switch) and \emph{switch-only} tasks (all switches with no move), totaling $57$ tasks. Test tasks are the remaining move and switch compositions. 
This setup directly evaluates whether a model that learned ``composition'' in a training setting (here, composition over moves) can transfer the ``composition'' ability to an unseen composition at test time (here, composition of moves and switches).

\textbf{Main observation.} At sample level, \ourTTT without model training solves \(0\%\) of composed test tasks, while \defaultTTT reaches near-perfect performance (\(\approx 99.95\%\)) due to full backbone adaptation on easy tasks. 
However, nearest-neighbor analyses in embedding space show the opposite trend for implicit rule induction: \ourTTT embeddings consistently retrieve the correct atomic constituents (move and switch) of the test compositions, whereas \defaultTTT largely fails to do so.

\textbf{Top-\(k\) dual-components retrieval.}
To quantify this effect, for each composed test task we check whether \emph{both} correct train move (without switch) and train switch (without move) are present in top-\(k\) nearest neighbors. As shown in Table~\ref{tab:move_switch}, \ourTTT reaches \(100\%\) by \(k=10\) (even by $k=7$), while \defaultTTT remains near zero for a wide range of \(k\) and only increases at very large \(k\). This indicates that \ourTTT preserves strong semantic locality for compositional rule components, whereas \defaultTTT does not. In addition, for each composed test task, we also measured: (i) whether the closest ``switch'' train neighbor is the correct switch, and (ii) whether the closest ``move'' train neighbor is the correct move. \ourTTT obtains \(100\%\) and \(99.5\%\), respectively; \defaultTTT obtains \(13.3\%\) (near random chance \(12.5\%\)) and \(40\%\).

\begin{table}[t]
\centering
\small
\caption{Top-\(k\) retrieval of both correct constituents (train move and train switch) for composed test tasks. Note that the entire dataset consists of k=57 tasks.}
\label{tab:move_switch}
\begin{tabular}{lcccccccc}
\toprule
TTT method & \(k=2\) & \(k=3\) & \(k=4\) & \(k=5\) & \(k=10\) & \(k=30\) & \(k=50\) & \(k=56\) \\
\midrule
\ourTTT     & 82.81\%  & 91.41\%  & 95.31\%  & 98.18\%  & 100.00\% & 100.00\% & 100.00\% & 100.00\% \\
\defaultTTT & 0.00\%   & 0.00\%   & 0.00\%   & 0.00\%   & 0.00\%   & 1.30\%   & 35.94\%  & 93.49\% \\
\bottomrule
\end{tabular}
\end{table}

\textbf{Interpretation.} These results support a dissociation between \emph{rule induction} and \emph{rule execution}: \ourTTT captures the intended compositional semantics in embedding space, but fails at execution because the frozen backbone was never trained to jointly realize move and switch transformations. This suggests that the bottleneck is primarily executability, not task-level rule identification. On the other hand, \defaultTTT unsurprisingly allows to solve the tasks, but at the cost of losing rule induction capabilities.





\end{document}